\documentclass[10pt,twocolumn,letterpaper]{article}

\usepackage{cvpr}      

\newcommand\blfootnote[1]{%
  \begingroup
  \renewcommand\thefootnote{}\footnote{#1}%
  \addtocounter{footnote}{-1}%
  \endgroup
}

\newcommand{\bx}{\mathbf{x}}
\newcommand{\bc}{\mathbf{c}}
\newcommand{\bp}{\mathbf{p}}
\newcommand{\bq}{\mathbf{q}}
\newcommand{\bs}{\mathbf{s}}

\newcommand{\bff}{\mathbf{f}}
\newcommand{\bE}{\mathbf{E}}
\newcommand{\bI}{\mathbf{I}}
\newcommand{\bX}{\mathbf{X}}
\newcommand{\bW}{\mathbf{W}}
\newcommand{\bA}{\mathbf{A}}
\newcommand{\bB}{\mathbf{B}}
\newcommand{\bepsilon}{\mathbf{\epsilon}}
\newcommand{\vae}{\mathcal{E}_{VAE}} 
\newcommand{\ani}{\mathcal{A}} 
\newcommand{\real}{\mathbb{R}}
\newcommand{\lpf}{\mathcal{L}}
\newcommand{\dct}{\mathcal{DCT}}
\newcommand{\idct}{\mathcal{IDCT}}

\newcommand{\caseB}{\mathcal{B}}
\newcommand{\caseF}{\mathcal{F}}
\newcommand{\caseG}{\mathcal{G}}
\newcommand{\caseH}{\mathcal{H}}
\newcommand{\caseP}{\mathcal{P}}
\newcommand{\caseR}{\mathcal{R}}
\newcommand{\caseS}{\mathcal{S}}
\newcommand{\caseW}{\mathcal{W}}
\newcommand{\caseC}{\mathcal{C}}

\usepackage{graphicx}
\usepackage{amsmath}
\usepackage{amssymb}
\usepackage{booktabs}
\usepackage[utf8]{inputenc}
\usepackage{algorithm, algpseudocode}
\usepackage{animate}
\usepackage{multirow}
\usepackage{cuted}  
\usepackage{capt-of}  
\usepackage{graphicx} 
\usepackage{makecell} 
\usepackage{graphicx}
\usepackage{rotating}

\definecolor{cvprblue}{rgb}{0.21,0.49,0.74}
\usepackage[pagebackref,breaklinks,colorlinks,allcolors=cvprblue]{hyperref}

\title{Few-Shot-Based Modular Image-to-Video Adapter for Diffusion Models}

\author{
Zhenhao Li\textsuperscript{*1} \and Shaohan Yi\textsuperscript{*2} \and Zheng Liu\textsuperscript{\dag1} \and Leonartinus Gao\textsuperscript{\dag3} \and Minh Ngoc Le\textsuperscript{\dag4} \and Ambrose Ling\textsuperscript{4} \and Zhuoran Wang\textsuperscript{4}
\and Md Amirul Islam\textsuperscript{1}
\and Zhixiang Chi\textsuperscript{1}
\and Yuanhao Yu\textsuperscript{1} 
\and
\\
\vspace{-5mm}
\small
\textsuperscript{1}Huawei Technologies Canada
\textsuperscript{2}University of Waterloo
\textsuperscript{3}University of British Columbia
\textsuperscript{4}University of Toronto
}

\begin{document}
\maketitle
\blfootnote{\textsuperscript{*\dag}Equal contribution. 
Shaohan, Leonartinus, Minh, Ambrose and Zhuoran did the work during internships at Huawei.}





\begin{abstract}
Diffusion models (DMs) have recently achieved impressive photorealism in image and video generation.
However, their application to image animation remains limited, even when trained on large-scale datasets.
Two primary challenges contribute to this:
the high dimensionality of video signals leads to a scarcity of training data, causing DMs to favor memorization over prompt compliance when generating motion;
moreover, DMs struggle to generalize to novel motion patterns not present in the training set, and fine-tuning them to learn such patterns, especially using limited training data, is still under-explored.
To address these limitations, we propose Modular Image-to-Video Adapter (MIVA), a lightweight sub-network attachable to a pre-trained DM, each designed to capture a single motion pattern and scalable via parallelization.
MIVAs can be efficiently trained on approximately ten samples using a single consumer-grade GPU.
At inference time, users can specify motion by selecting one or multiple MIVAs, eliminating the need for prompt engineering. Extensive experiments demonstrate that MIVA enables more precise motion control while maintaining, or even surpassing, the generation quality of models trained on significantly larger datasets.
\end{abstract}

\section{Introduction}
\label{sec:intro}



Image animation has been a consistently evolving technology for years alongside the growth of mass media industries.
Applications such as Live2D have made animation technology easily accessible to the public, allowing amateur users to create animations from any static image, also known as image-to-video (I2V) generation. 
However, traditional I2V methods often require significant technical expertise and manual labor, which limits their widespread use.
Recently, diffusion models (DMs) have emerged as potential game changers for animation technology,
showcasing unprecedented capabilities in generating videos.
Most DM-based frameworks utilize both a reference image and a text prompt to guide motion synthesis, making animation more accessible to casual users than conventional techniques.

\begin{figure}[t]
\centering
\setlength{\tabcolsep}{0pt} 
\begin{tabular}{p{0.33\linewidth}p{0.33\linewidth}p{0.33\linewidth}c}

\parbox{\linewidth}{
        \centering
        \includegraphics[width=0.48\linewidth]{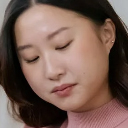}%
        \includegraphics[width=0.48\linewidth]{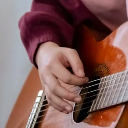}%
    }  &
\parbox{\linewidth}{
        \centering
        \includegraphics[width=0.48\linewidth]{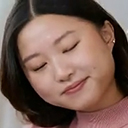}%
        \includegraphics[width=0.48\linewidth]{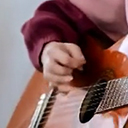}%
    }  &
\parbox{\linewidth}{
        \centering
        \includegraphics[width=0.48\linewidth]{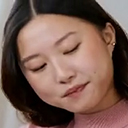}%
        \includegraphics[width=0.48\linewidth]{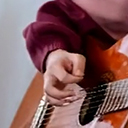}%
    } &
\\
\footnotesize \centering Input & 
\footnotesize \centering Wan: ``The woman looks to the ceiling.'' &
\footnotesize \centering Wan: ``... smiles, holding the guitar.'' &
\\

\parbox{\linewidth}{
        \centering
        \includegraphics[width=0.48\linewidth]{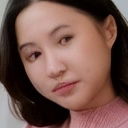}%
        \includegraphics[width=0.48\linewidth]{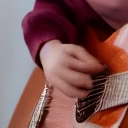}%
    }  &
\parbox{\linewidth}{
        \centering
        \includegraphics[width=0.48\linewidth]{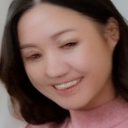}%
        \includegraphics[width=0.48\linewidth]{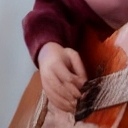}%
    }  &
\parbox{\linewidth}{
        \centering
        \includegraphics[width=0.48\linewidth]{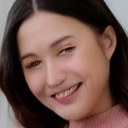}%
        \includegraphics[width=0.48\linewidth]{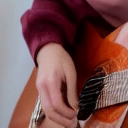}
    }
\\
\footnotesize \centering MIVA: guitar playing & 
\footnotesize \centering MIVA: turn to smile &
\footnotesize \centering 2 MIVAs combined &
\\

\end{tabular}


\small
\caption{Animation result (last frame) by Wan2.1-14B-I2V and MIVA, given an image of a woman holding a guitar.
Wan ignores user prompt, always producing guitar-playing animation. 
In contrast, MIVAs display higher user controllability. We crop the face and hand patches for better view. 
   }
\label{fig:intro}

\end{figure}

While powerful, DMs face critical challenges that hinder their effectiveness.
To begin with, the complexity of real-world motion and the high-dimensional nature of video signals result in the inherent sparsity of training data across the text-image-video training space.
This scarcity can lead models to overfit, often replicating training samples rather than adapting to user prompts \cite{replication}.
For instance, as shown in Fig.~\ref{fig:intro}, the DM consistently generates guitar-playing animations regardless of the input prompt, though well trained on large-scale datasets.
This replication issue greatly undermines user control and customization.
Moreover,
although an I2V DM trained on massive data is expected to generalize well, it often fails with niche or specific motion patterns (see Fig.~\ref{fig:single}). Users may want to fine-tune the I2V DM to optimize the synthesis of such motion with minimal cost, but this problem remains largely under-explored.

In this paper,
we propose a distinct paradigm, dubbed \textit{modular I2V}, to instantiate DM-based I2V generation.
Specifically, given a base DM, we handle a motion pattern by a dedicated \textit{adapter}, a lightweight sub-network attached to the base DM, termed a Modular I2V Adapter (MIVA). 
A MIVA takes surprisingly low data collection and training cost. 
During inference, 
users can select and employ one or multiple MIVAs, with optional respective weights, to specify their desired motion. 
Furthermore, similar to the vision of modular customization for image synthesis \cite{oa}, modular I2V enables users to share their own trained adapters, and/or deploy the adapters from others with ease.


For the design of MIVA, we take inspiration from LAMP \cite{lamp}, a single-motion-pattern I2V framework. 
But LAMP as a model fine-tuned from image DM is not ideal for modular I2V, mainly because LAMP learns all temporal module parameters from scratch. This results in LAMP's cumbersome temporal modules (exceeding one third of the base DM), along with instability in training and poor generalization ability in inference. 
Nor does LAMP address the multi-motion-pattern cases.
To overcome these limitations,
we adopt a video generation DM as the foundation, leveraging its inherent motion priors.
Each MIVA is implemented as parameter-efficient, learnable weights parallel to each attention block in the base DM, comprising approximately 3\% of the base model's parameters.
By associating a MIVA with a learnable implicit embedding, we eliminate the need for text prompt engineering, making the selection and weighting of MIVAs the only control mechanism for users.
Additionally, we introduce a subject-mask-based extension:
given the instance segmentation mask of the motion subject (automatically derived from the input image), MIVA jointly generates frames and a subject mask sequence, and meanwhile leverages the mask sequence to refine the generation of frames, enhancing motion realism and improving robustness against imperfections in the training data.


We conduct extensive experiments on MIVA, demonstrating its superior performance on both single-motion-pattern animation and more complex multi-motion-pattern animation, achieved by parallelizing multiple MIVAs.
Our method is 
validated by quality assessment algorithms
as well as comprehensive user studies, highlighting its effectiveness and robustness.
The key contributions of this work are as follows:
1) we introduce a modular I2V paradigm that decomposes the image animation task into atomic motion components, each addressed by a dedicated MIVA;
2) we propose a parameter-efficient MIVA design that supports few-shot training and enables streamlined deployment at inference;
3) we explore a subject-mask-based enhancement strategy whereby mask sequences are generated in tandem with video frames and used to refine frame synthesis.
\section{Related Works}
\label{s:related}

\subsection{Image animation (I2V) DMs}
\label{s:i2v}
Image animation is a video generation task 
aiming to produce realistic motion from a static image input, often one of the final video frames.
An intuitive idea is to impose an image-conditioning mechanism on a pre-trained T2V model, but the design of such mechanisms is nontrivial.
Options vary from 1) integrating the image into the noise signal via forward diffusion process \cite{cil}, concatenation channel-wise \cite{pia} or frame-wise \cite{lamp, aa, consisti2v, trip, cinemo}, 2) imposing classifier-free guidance (CFG)\cite{cfg} with a feature encoding the image \cite{svd, ip-adapter, videocrafter1} (\eg by a CLIP image encoder \cite{clip}), and 3) doing both \cite{videocomposer, dynamicrafter, i2v-adapter, i2vgen}.
It is disputable whether image-embedding-based CFG can benefit I2V generation performance.
\cite{animatelcm, cil, cinemo, consisti2v} criticize that CLIP image encoder focuses more on the high-level semantic information rather than the visual details, which are more important in the I2V problem.
\cite{dynamicrafter, ip-adapter} propose to learn a projection module upon the CLIP image embedding to introduce inductive bias towards visual details, but this requires training on large-scale datasets. 
\cite{cinemo, trip} guide the DM to learn the difference between each frame and the reference image to enhance temporal coherency, and some methods take additional control signals from users, such as the mask of the subject(s) to animate \cite{aa} and motion trajectories \cite{mofa, motioni2v}
, which are data-demanding as well.
We note some methods specializing in specific motion types, such as the oscillating motion of foreground objects \cite{gid}, motion of hair \cite{hair}, portrait \cite{liveportrait, xportrait} and body \cite{animateanyone, dreampose} \etc, while this paper is concerned with open-domain motion patterns.
We also highlight a related task known as motion transfer \cite{flexiact}, which involves generating a video of a specific subject mimicking the motion in a reference video. Unlike the general-purpose I2V problem studied in this paper, which is designed to generalize across different subjects, motion transfer typically necessitates subject-specific fine-tuning, limiting its scalability and adaptability.


\begin{figure*}[t]
  \centering
  \includegraphics[width=0.9\linewidth]{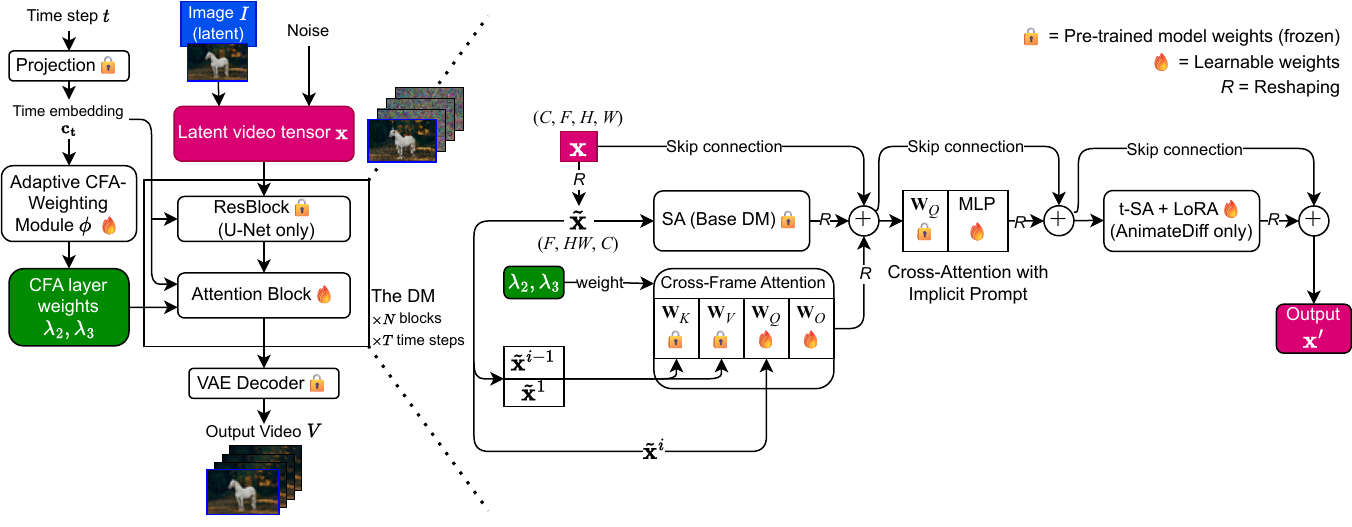}
   \caption{Architecture of MIVA. 
    For simplicity, we assume that all signals during the diffusion process are in the latent domain.
   }
   \label{fig:main}
\end{figure*}

\subsection{Parameter-Efficient Fine-Tuning (PEFT)}
\label{s:peft}
PEFT aims to adapt models by optimizing a small subset of parameters, significantly reducing computational and storage overhead.
In the DM context, existing PEFT methods vary in their optimization targets,
ranging from model parameters \cite{CustomDiffusion}, adapters \cite{lora, ip-adapter} to non-DM add-ons (\eg, text embeddings \cite{ti}). 
Among these, low-rank adaptation (LoRA) \cite{lora} stands out for its effectiveness and versatility, offering lightweight fine-tuning for any DM.
Its architecture allows adaptations to arbitrary linear layers, granting exceptional flexibility when customizing model behavior.
In practice, LoRA has enabled fine-grained personalization in image and video generation tasks, from specific subject and style \cite{paired} to motion \cite{animatediff}.
Moreover, its ability to generalize with minimal training data \cite{lora-dm} makes it especially compelling in few-shot scenarios.
These strengths motivate our exploration of LoRA in modular, personalized I2V generation, a space that remains under-explored, with
LAMP \cite{lamp} being the only prior solution to our knowledge.

\section{Method}
\label{s:method}

\subsection{Preliminary}
\label{s:dm}

Our approach is based on the latent diffusion model (LDM) framework \cite{ldm}.
Architecturally, most DMs adopt either a U-Net \cite{unet} or diffusion transformer (DiT) \cite{dit} backbone, both of which 
employ transformer blocks \cite{transformer} as their main building components. 
A transformer block includes up to two types of 
attention layers: self-attention (SA) and cross-attention (CA),
both governed by the attention function
$Att(Q, K, V) = \sigma \left(QK^T / \sqrt{d_K} \right) V$,
where $Q$, $K$, $V$ are query, key, and value matrices respectively, $d_K$ is the dimensionality of $K$ features, and $\sigma(\cdot)$ is the softmax function.
SA and CA differ in the setting of $Q$, $K$ and $V$.
Given a token sequence $\bff$ representing a series of video patches, and another sequence $\bc$ representing the text prompt, SA and CA have form
\begin{subequations} \label{eq:att}
\begin{align}
SA(\bff; \theta) &= 
Att(\bff \bW_Q, \bff \bW_K, \bff \bW_V) \bW_O, \label{eq:a}\\
CA(\bff, \bc; \theta) &= Att(\bff \bW_Q, \bc \bW_K, \bc \bW_V) \bW_O, \label{eq:b}
\end{align}
\end{subequations}
$\theta := \{\bW_{Q,K,V,O}\}$ being the projection matrices associated with the attention layer of interest.
SA models interactions between patches within $\bff$ (whose scope can range from a single frame to the full video depending on the DM's temporal design);
CA, by contrast, captures interactions between the video and external conditioning inputs.


\subsection{Problem Setting and Motivation}
\label{s:setting}

We adopt the problem setting introduced by LAMP \cite{lamp}, which trains a DM to animate an input image $I$ with a target motion pattern $\ani$,
using a small-scale dataset (size at $10^1$ scale) specific to the motion pattern. The output is a video $V=\{I^i | i= 1, \dots, F\}$ with $I^1 \equiv I$ serving as the initial frame.
While we build upon this setup, our work introduces two key modifications aimed at improving training stability and controllability, as detailed below.

\subsubsection{T2V DM as Base Model}

Unlike LAMP, which trains all temporal layers from scratch and exhibits notable instability across motion patterns due to overfitting on scarce data,
we leverage the temporal layers from pre-trained T2V DMs,
which can be easily fine-tuned \cite{animatediff},
to exploit the rich motion priors therein.
This enables substantial reduction in both training cost and risk of training instability.

\subsubsection{Modular I2V}
We study modular I2V, a more generalized image animation setting than LAMP: the desired animation $\ani^*$ can be interpreted as one or multiple atomic motion patterns $\{\ani_i\}$,
each corresponding to a different moving subject,
handled by a corresponding model $D_{\theta_i}$.
We further assume 
all models $\{\theta_i\}$ to be parallelizable, allowing them to be integrated into a single DM $D^*$ that animates the input image with $\ani^*$ end-to-end, unlike a LAMP model which cannot be applied in parallel with another.

To this end, we formulate $\theta_i$ as $\theta_i:=\{\theta_0, \theta_i^*\}$, $\theta_0$ being the fixed parameters from a common base DM and $\theta_i^*$ being the learnable parameters.
During inference, the operations involving $\theta_i^*$ are independent of $\theta_0$ but modify the results of certain intermediate layers,
functioning as an \textit{adapter} attached to the base model $D_{\theta_0}$.
Multiple adapters can be attached to $D_{\theta_0}$ to instantiate $D^*$,
merging the motion concepts from each adapter.
We refer to each $\theta_i^*$ as a Modular I2V Adapter (MIVA).

\subsection{Modular I2V Adapter (MIVA)}
\label{s:main}

As illustrated in Fig.~\ref{fig:main}, a MIVA consists of three specialized submodules, respectively situated at SA, CA, and temporal SA (t-SA) layers.
Given that several video generation methods \cite{videoldm, animatediff} demonstrate good performance without altering ResBlocks,
these submodules are integrated exclusively with attention layers, leaving the ResBlocks untouched.
This design choice also ensures compatibility with ResBlock-free DMs,
in particular DiTs.
Next, we detail the architecture and functionality of each submodule, aligned with its respective attention layer.

\subsubsection{Cross-frame Attention (CFA) Layers}
\label{s:sa}


Cross-frame correlation plays a crucial role in I2V generation.
During synthesis of frame $I^i$, its relationships with $I^1$ and the preceding frame $I^{i-1}$ should be emphasized to ensure appearance consistency and temporal smoothness.
Inspired by Tune-A-Video \cite{tav}, we augment each SA layer with two CFA layers, designed to model the 
$I^i$-$I^1$ and $I^i$-$I^{i-1}$ dependencies, respectively.
Each CFA layer between $I^i$ and a reference frame $I^j$ ($j \in \{1, i-1\}$) is defined as $CFA_{i, j} := Att(\bff^i\bW_Q, \bff^j \bW_K, \bff^j \bW_V) \bW_O$, $\bff^i$ being the token sequence representing $I^i$.
The augmented SA block is formulated as a weighted combination:
\begin{equation}
\label{eq:sa}
SA^*(\bff^i; \bff^1, \bff^{i-1}) = \lambda_1 SA(\bff^i) + \lambda_2 CFA_{i, 1} + \lambda_3 CFA_{i, i-1}.
\end{equation}
We optimize $\bW_Q$ and $\bW_O$ in each CFA layer, initializing them with the base model weights and zeros, respectively;
$\bW_K$ and $\bW_V$ are based on the corresponding pre-trained weights and kept frozen.

Previous studies treat $\{ \lambda_{1,2,3} \}$ as fixed hyperparameters.
Options include all $1$s
in Tune-A-Video, $(0.7, 0.15, 0.15)$ in PoseAnimate \cite{poseanimate} and $(1,1,0)$ in I2V-Adapter \cite{i2v-adapter}.
However, we find the weighting of CFA layers non-trivial due the diversity of motion patterns.
In general, faster or erratic motion (\eg, raindrops) requires higher $\lambda_2$ to enforce consistency with $I^1$, which anchors global appearance.
Meanwhile, slower or subtler motion (\eg, humans, animals) favors a higher $\lambda_3$ to maintain continuity with the previous frame.
The $\lambda$-assignment problem is even more intricate
when considering the diffusion time step:
given the insight that different visual contents are formed at different time steps \cite{directed}, the optimal $\lambda$ configuration can shift over time.

To alleviate the burden of manual $\lambda$-assignment,
we propose an \textit{adaptive weighting module} $\phi$, attached to each SA layer. This module is defined as $\phi(\bc_t; \bW_\phi) = \sigma (\mathrm{SiLU}(\bc_t) \bW_\phi)$ where $\bc_t \in \real^{d_t}$ is the time step embedding and $\bW_\phi \in \real^{d_t \times 2}$ is a learnable projection.
This module dynamically outputs a 2-dimensional weighting vector, which is assigned to $(\lambda_2, \lambda_3)$, while $\lambda_1$ is fixed to $1$.
We demonstrate the impact of $\phi$ in the ablation study section.

\subsubsection{CA Layers with Implicit Prompt}


Although our modular I2V framework does not rely on text prompts from users, we can still leverage the priors within the CA layers of the base model. 
Since each MIVA specializes in a single motion type, we associate it with a fixed prompt that semantically represents its target motion pattern.
We also assign this prompt uniformly to all training videos exhibiting the target motion. Appearance information is omitted from the prompt, as it is already provided by $I$.
Rather than using natural language, we encode this motion-specific prompt as a learnable embedding $\bc$, which we refer to as an \textit{implicit prompt}, eliminating the need for manual prompt engineering.


According to Eq. \eqref{eq:b}, a CA layer can be written as
$Att(\bff \bW_Q, \bc \bW_K, \bc \bW_V) \bW_O = \sigma ( \frac{\bff \bW_Q \bW_K^T \bc^T}{\sqrt{d}} ) \bc \bW_V \bW_O.$
We preserve $\bW_Q$, which encodes the video signal, to retain the semantic prior from the base model, while fine-tuning the remaining components.
The CA operation can thus be reformulated as
$\sigma(\bff \bW_Q \mathbf{A}) \mathbf{B}$, where $\mathbf{A}=\frac{1}{\sqrt{d}}\bW_K^T \bc^T$ and $\mathbf{B}=\bc \bW_V \bW_O$, equivalent to applying a 2-layer perceptron (MLP) following the fixed projection $\bW_Q$.
As a result, we substitute the original CA layer with the above MLP structure, greatly reducing the module size to LoRA-level without compromising the module's conditioning capacity.

\subsubsection{Temporal SA Adapter}
Following AnimateDiff \cite{animatediff}, 
we fine-tune each projection matrix within the t-SA layers using corresponding LoRAs, in order to adapt temporal modules to novel motion patterns.

\begin{figure}[t]
  \centering
  \includegraphics[width=0.95\linewidth]{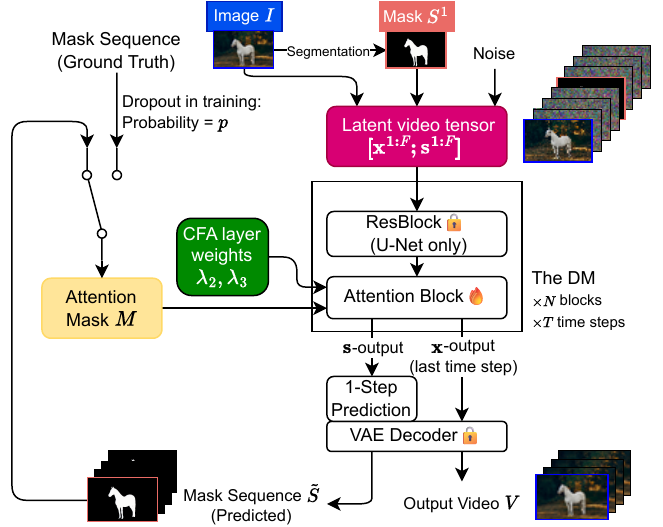}
   \caption{Masked MIVA (M-MIVA) generates not only video frames but also the subject mask sequence, which is then modulated into the attention mask $M$ that guides the subsequent time steps in the diffusion process.
   }
   \label{fig:mask}
\end{figure}

\subsection{Masked MIVA (M-MIVA)}

The lightweight design of MIVA enables training in few-shot scenarios without requiring manual annotations.
However, in the absence of constraints, MIVA may indiscriminately learn from all moving pixels in the training samples.
This makes it susceptible to data imperfections, such as 
incidental motion from unrelated subjects, and biases inherent in limited samples, often leading to visible artifacts
(see Fig.~\ref{fig:ablation_mask}).
To improve robustness against data imperfections,
we incorporate additional cues that help the DM focus on the subject of interest.
Inspired by Through-The-Mask \cite{through-the-mask},
we utilize subject-specific mask sequences to distinguish foreground from background within each frame, effectively guiding the model's attention.
We employ an off-the-shelf detect-and-segment pipeline \cite{groundingdino, sam2},
which requires no additional user input,
to extract the subject masks from sample videos during training and from $I$ during inference.

Without disrupting the modular I2V framework, we propose 
Masked MIVA (M-MIVA),
which jointly generates both the video $V$ and the mask sequence $S$, while simultaneously using $S$ to enhance the quality of $V$.
Unlike Through-The-Mask that employs two separate DMs to handle video and mask modalities, MIVA achieves this dual output within a single diffusion process.
This is enabled by our observation that intermediate mask generation can effectively guide video synthesis via attention masking mechanisms.

As illustrated in Fig.~\ref{fig:mask}, M-MIVA operates on a video tensor
$[\bx^{1:F}; \bs^{1:F}]$, 
formed by concatenating the generated frame sequence $\{ \bx^{1:F}\}$ and its corresponding mask sequence $\{ \bs^{1:F}\}$.
The initial frame and mask,
$\bx^1$ and $\bs^1$,
are set to $I$ and its subject mask in the latent domain.
M-MIVA can be trained using the same denoising objective as the base DM,
with key modifications to the attention layers that enable interaction between video and mask modalities.
Specifically,
we introduce a parallel mask-modality attention stream to each SA layer, composed of independently parameterized CFA layers. Each frame-mask CFA pair is identically initialized to preserve initial symmetry.
In addition, the mask sequence is transformed into attention masks that guide the attention computation in the video-modality stream, as described in the following subsection.

\subsubsection{Attention Mask}


An attention mask $M$ can be used to bias the attention weights between query-key token pairs, yielding the modified attention expression $\sigma \left( \frac{QK^T}{\sqrt{d}} + M \right)V$.
In I2V context,
we aim to promote attention between video tokens that belong to the same semantic region (foreground or background), and suppress attention across disparate regions.
This encourages focus on the most relevant areas during motion learning.
Denote by $\bx^i_{\bp}$ the video token at spatial location $\bp$ in $I^i$, and by $S^i_{\bp}$
the token's subject confidence score, \ie, the value from $S^i$.
Then, for any token pair $(\bp,i)$ and $(\bq,j)$, we define the attention mask entry $M^{(i, j)}_{(\bp,\bq)}$ as
\begin{equation}
\label{eq:mask}
M^{(i, j)}_{(\bp,\bq)} = \log \left(S^i_\bp S^j_\bq + (1-S^i_\bp)(1-S^j_\bq) + \epsilon \right),
\end{equation}
where $\epsilon$ is a small constant. $M$ takes 
large negative values for cross-region token pairs, thereby suppressing attention.


Since the mask sequence $S$ is not readily available during intermediate steps of the diffusion process, we approximate $S$ through a one-step prediction strategy. Specifically, at time step $t$, we use the current $\{ \bs_t^{1:F}\}$ to predict a proxy mask $\tilde S$.
Assuming the DM is a denoising LDM $\bepsilon_\theta(\bs_t, t)$ with noise schedule $(\alpha_t, \sigma_t)$, the approximation 
$\tilde S = D_{VAE}\left(\frac{1}{\alpha_t} \left(\bs_t - \sigma_t \bepsilon_\theta(\bs_t, t)\right)\right)$,
where $D_{VAE}$ is the decoder of the pre-trained variational autoencoder (VAE).

\begin{figure}
    \centering
    \parbox{0.22\linewidth}{
        \centering
        \includegraphics[width=\linewidth]{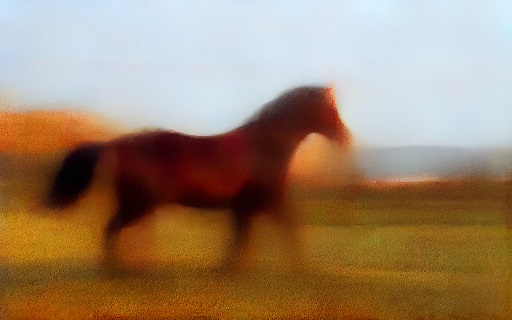}
        
        \includegraphics[width=\linewidth]{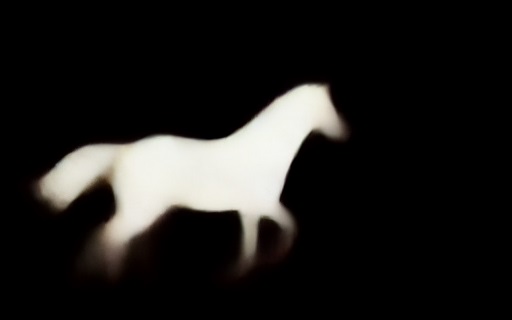}
        
        {\small 1000}
    }
    \parbox{0.22\linewidth}{
        \centering
        \includegraphics[width=\linewidth]{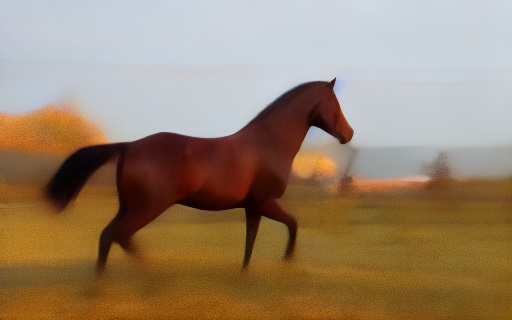}
        
        \includegraphics[width=\linewidth]{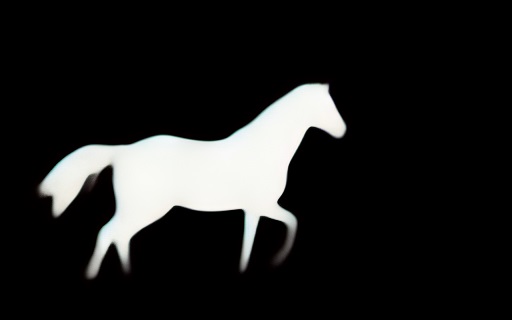}
        
        {\small 750}
    }
    \parbox{0.22\linewidth}{
        \centering
        \includegraphics[width=\linewidth]{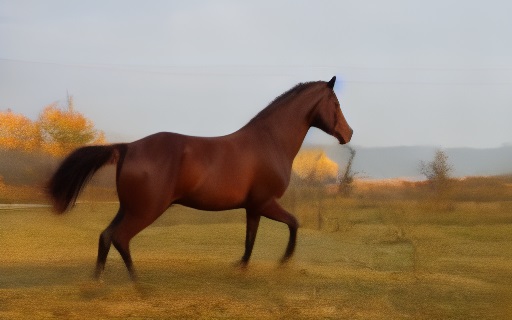}
        
        \includegraphics[width=\linewidth]{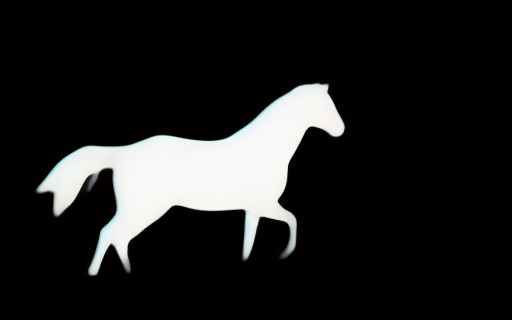}
        
        {\small 500}
    }
    \parbox{0.22\linewidth}{
        \centering
        \includegraphics[width=\linewidth]{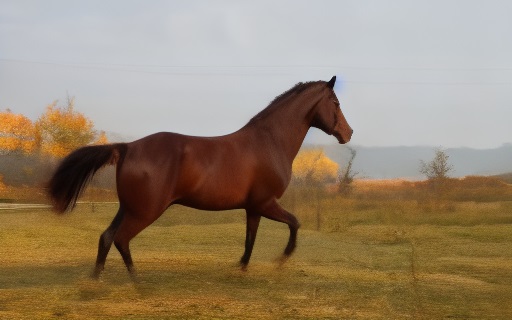}
        
        \includegraphics[width=\linewidth]{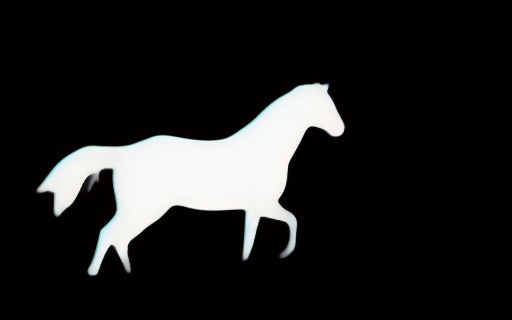}
        
        {\small 250}
    }
    \caption{One-step prediction of the last frame (top) and its subject mask (bottom) at different time steps.}
    \label{fig:1step}
\end{figure}

\subsubsection{Dropout Training}
During early training,
the predicted mask sequence $\tilde S$ is often too noisy to serve as a reliable basis for attention modulation.
To mitigate this, we employ a dropout-based strategy:
with probability $p$,
we replace $\tilde S$ with the ground truth $\{S_0^{1:F}\}$,
where $p$ follows a predefined decreasing schedule.
This mechanism stabilizes early attention behavior and prevents the learning process from being corrupted by inaccurate mask predictions.

\begin{table*}[t]
    \centering
    \resizebox{1\textwidth}{!}{
    \renewcommand{\arraystretch}{1} 
    \begin{tabular}{c|l|cccccccc}
        \toprule
        \makecell{Training \\ Data} & Method & \makecell{Subject \\ Consistency $\uparrow$} & \makecell{Background \\ Consistency $\uparrow$} & \makecell{Motion \\ Smoothness $\uparrow$} & \makecell{Temporal \\ Flickering $\uparrow$} & \makecell{Average \\ Flow $\uparrow$} & \makecell{Aesthetic \\ Quality Loss $\downarrow$} & \makecell{Image \\ Quality Loss $\downarrow$} & \makecell{User \\ Preference (\%) $\uparrow$} \\
        \midrule

        \multirow{5}{*}{\makecell{Large-\\scale}} &
        SVD \cite{svd} 
        & $93.7 ~ \vert ~ 95.2$ 
        & $96.3 ~ \vert ~ 96.6$ 
        & $98.4 ~ \vert ~ 96.9$ 
        & $96.7 ~ \vert ~ 93.3$ 
        & $\underline{5.82} ~ \vert ~ \mathbf{5.55}$
        & $8.16 ~ \vert ~ 7.79$ 
        & $3.70 ~ \vert ~ 3.80$ 
        & $3.3 ~ \vert ~ 5.2$ \\ 
               
        & I2VGen-XL \cite{i2vgen} 
        & $94.5 ~ \vert ~ 92.9$ 
        & $97.5 ~ \vert ~ 96.1$ 
        & $98.4 ~ \vert ~ 98.1$ 
        & $96.7 ~ \vert ~ 96.0$ 
        & $4.08 ~ \vert ~ \underline{4.64}$ 
        & $\underline{4.52} ~ \vert ~ 5.24$ 
        & $3.64 ~ \vert ~ 3.19$ 
        & $\underline{17.6} ~ \vert ~ \underline{19.9}$ \\

        & DynamiCrafter \cite{dynamicrafter} 
        & $95.5 ~ \vert ~ 97.0$ 
        & $97.5 ~ \vert ~ 97.7$ 
        & $97.4 ~ \vert ~ 98.3$ 
        & $94.1 ~ \vert ~ 95.7$ 
        & $\mathbf{5.93} ~ \vert ~ 3.47$ 
        & $\mathbf{1.21} ~ \vert ~ \mathbf{1.34}$ 
        & $\underline{0.88} ~ \vert ~ \mathbf{0.48}$ 
        & $13.4 ~ \vert ~ 7.2$ \\
        
        & I2V-Adapter \cite{i2v-adapter} 
        & $95.8 ~ \vert ~ 95.7$ 
        & $97.6 ~ \vert ~ 97.0$ 
        & $97.7 ~ \vert ~ 98.0$ 
        & $96.6 ~ \vert ~ 96.7$ 
        & $1.77 ~ \vert ~ 1.61$ 
        & $4.83 ~ \vert ~ 6.04$ 
        & $3.98 ~ \vert ~ 6.44$ 
        & $5.4 ~ \vert ~ 6.9$ \\
        
        & Cinemo \cite{cinemo} 
        & $\underline{95.9} ~ \vert ~ \mathbf{98.3}$ 
        & $\mathbf{98.1} ~ \vert ~ \mathbf{98.8}$ 
        & $\mathbf{99.0} ~ \vert ~ \mathbf{99.4}$ 
        & $\mathbf{98.2} ~ \vert ~ \mathbf{99.0}$ 
        & $0.50 ~ \vert ~ 0.42$ 
        & $6.75 ~ \vert ~ 6.70$ 
        & $6.43 ~ \vert ~ 6.98$ 
        & $16.7 ~ \vert ~ 18.7$ \\
        
        \midrule
        \multirow{3}{*}{Few-shot} &
        LAMP \cite{lamp} 
        & $92.9 ~ \vert$ N/A 
        & $95.8 ~ \vert$ N/A 
        & $97.1 ~ \vert$ N/A 
        & $97.7 ~ \vert$ N/A 
        & $1.65 ~ \vert$ N/A
        & $7.25 ~ \vert$ N/A 
        & $1.29 ~ \vert$ N/A 
        & $9.1 ~ \vert$ N/A  \\
        

        & MIVA 
        & $\mathbf{96.6} ~ \vert ~ \underline{97.7}$
        & $\underline{98.0} ~ \vert ~ \underline{98.1}$ 
        & $\underline{98.8} ~ \vert ~ \underline{99.0}$ 
        & $\underline{97.9} ~ \vert ~ \underline{98.2}$ 
        & $1.40 ~ \vert ~ 1.20$ 
        & $4.84 ~ \vert ~ \underline{4.89}$  
        & $\mathbf{0.52} ~ \vert ~ \underline{0.53}$  
        & $\mathbf{34.5} ~ \vert ~ \mathbf{42.1}$ \\

        & MIVA w/o CFA weighting module $\phi$ 
        & $93.8 ~ \vert$ N/A
        & $95.5 ~ \vert$ N/A 
        & $98.3 ~ \vert$ N/A 
        & $97.2 ~ \vert$ N/A 
        & $1.75 ~ \vert$ N/A
        & $7.00 ~ \vert$ N/A
        & $3.75 ~ \vert$ N/A  
        & N/A \\

        \bottomrule
    \end{tabular}
    \renewcommand{\arraystretch}{1} 
     }
     \caption{Quantitative comparison on our single-motion-pattern (left) and multi-motion-pattern (right) benchmark datasets. The best number is highlighted in \textbf{bold}, and the second best is \underline{underlined}.
    } 
    \label{tab:main}
\end{table*}
\begin{figure*}[!ht]
\centering
\setlength{\tabcolsep}{0pt} 
\begin{tabular}{c*{10}{c}}

\raisebox{9pt}{\rotatebox[origin=l]{90}{\small $\caseB$}}\hspace{4pt} &
\includegraphics[width=0.10\linewidth]{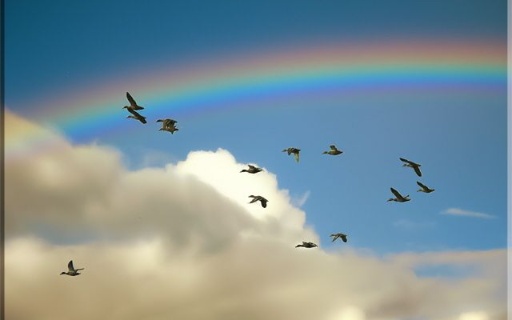}\hspace{2pt} &
\includegraphics[width=0.10\linewidth]{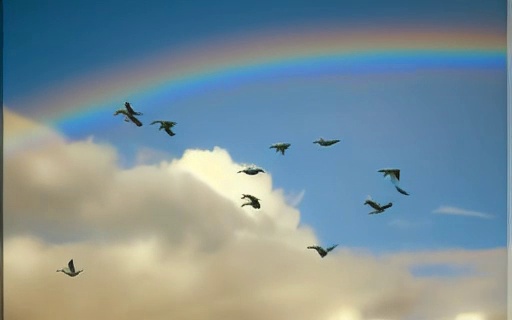} &
\includegraphics[width=0.10\linewidth]{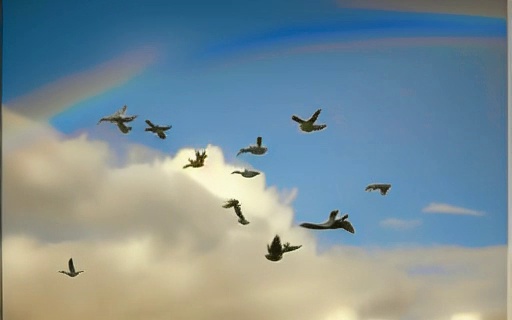}\hspace{2pt} &
\includegraphics[width=0.10\linewidth]{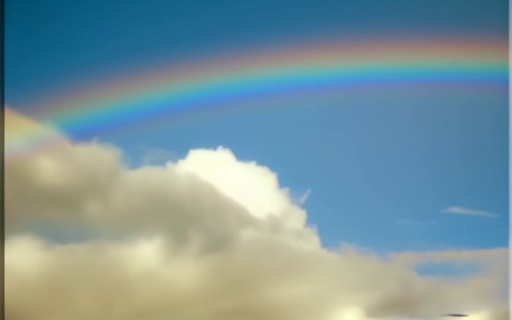} &
\includegraphics[width=0.10\linewidth]{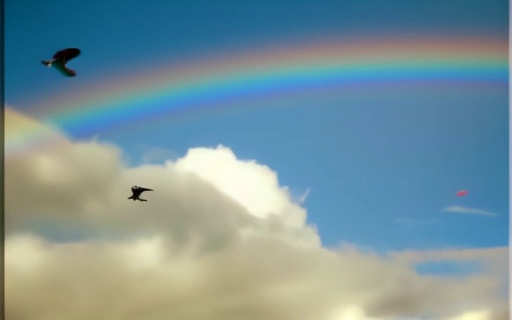}\hspace{2pt} &
\includegraphics[width=0.10\linewidth]{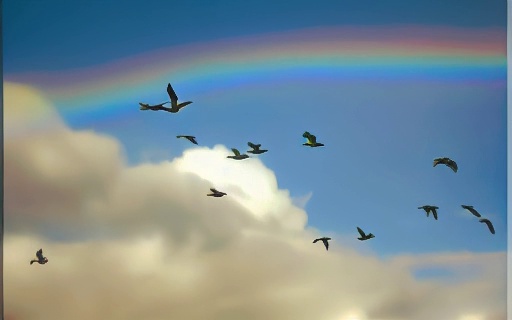} &
\includegraphics[width=0.10\linewidth]{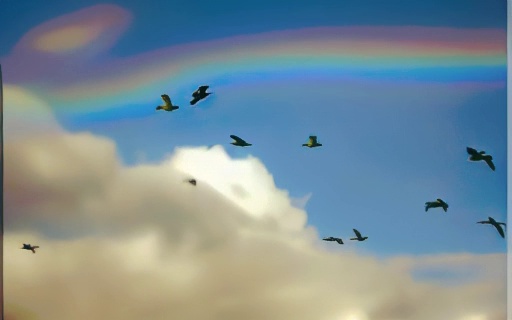}\hspace{2pt} &
\includegraphics[width=0.10\linewidth]{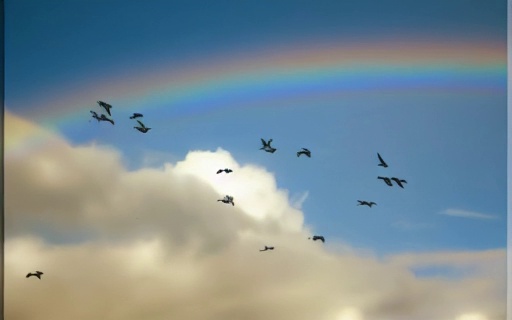} & 
\includegraphics[width=0.10\linewidth]{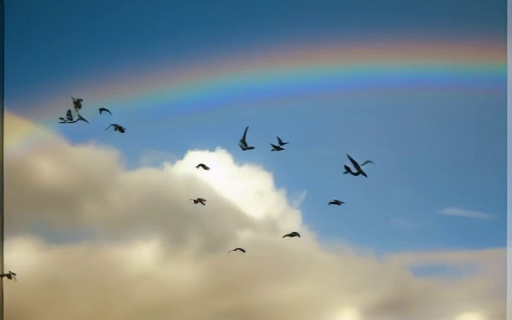}\hspace{2pt} & \\

\raisebox{9pt}{\rotatebox[origin=l]{90}{\small $\caseF$}}\hspace{4pt} &
\includegraphics[width=0.10\linewidth]{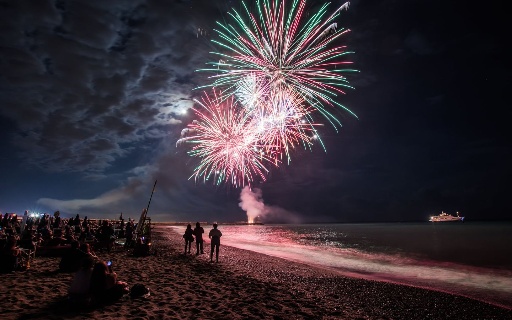}\hspace{2pt} &
\includegraphics[width=0.10\linewidth]{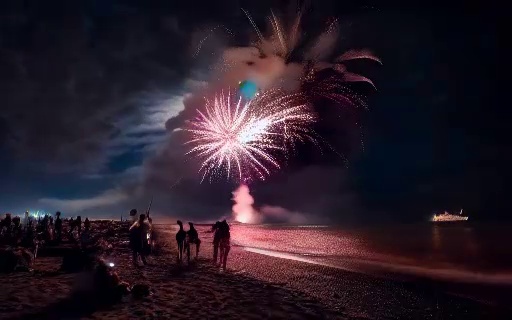} &
\includegraphics[width=0.10\linewidth]{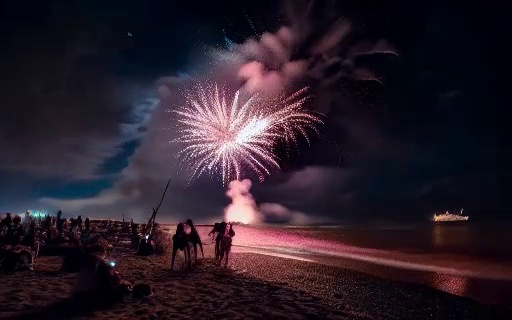}\hspace{2pt} &
\includegraphics[width=0.10\linewidth]{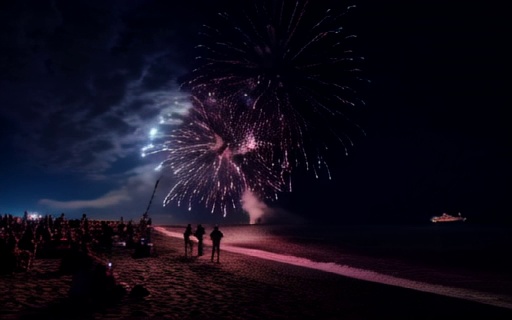} &
\includegraphics[width=0.10\linewidth]{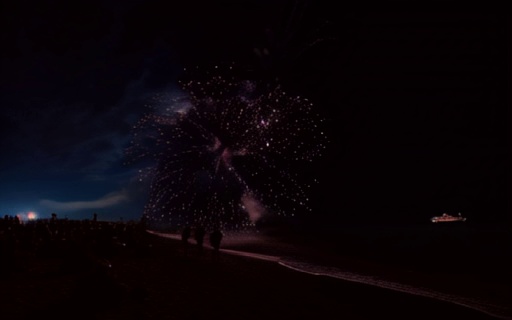}\hspace{2pt} &
\includegraphics[width=0.10\linewidth]{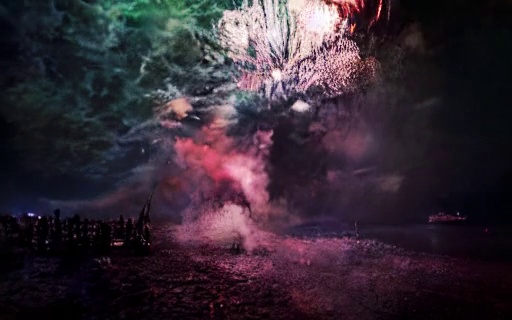} &
\includegraphics[width=0.10\linewidth]{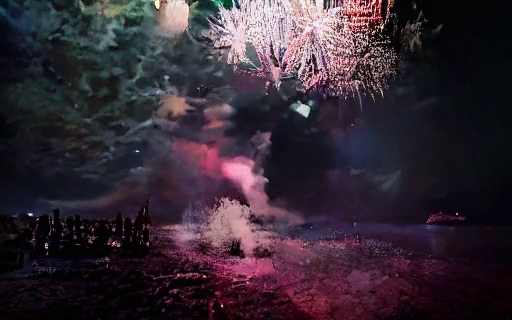}\hspace{2pt} &
\includegraphics[width=0.10\linewidth]{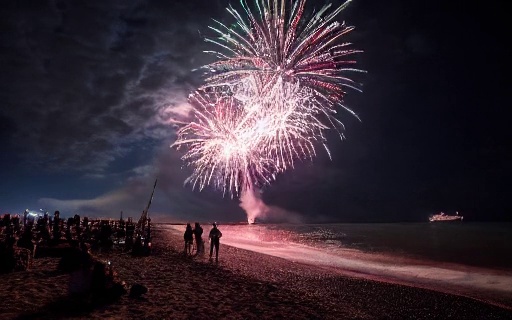} & 
\includegraphics[width=0.10\linewidth]{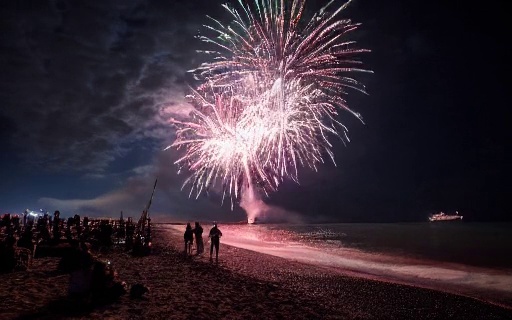}\hspace{2pt} & \\

\raisebox{9pt}{\rotatebox[origin=l]{90}{\small $\caseG$}}\hspace{4pt} &
\includegraphics[width=0.10\linewidth]{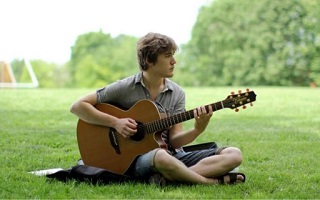}\hspace{2pt} &
\includegraphics[width=0.10\linewidth]{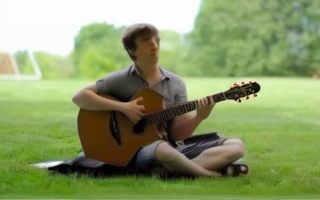} &
\includegraphics[width=0.10\linewidth]{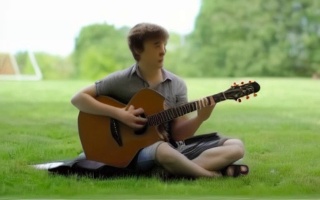}\hspace{2pt} &
\includegraphics[width=0.10\linewidth]{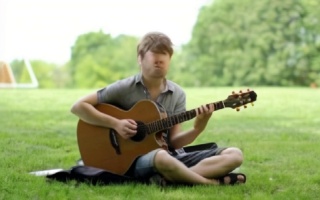} &
\includegraphics[width=0.10\linewidth]{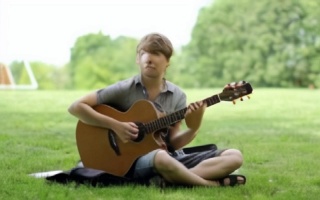}\hspace{2pt} &
\includegraphics[width=0.10\linewidth]{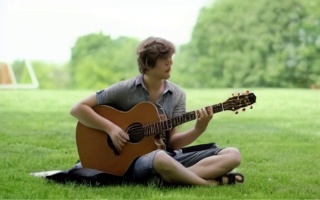} &
\includegraphics[width=0.10\linewidth]{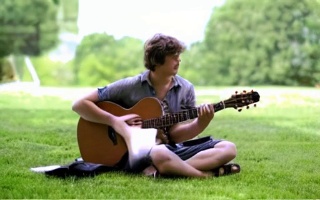}\hspace{2pt} &
\includegraphics[width=0.10\linewidth]{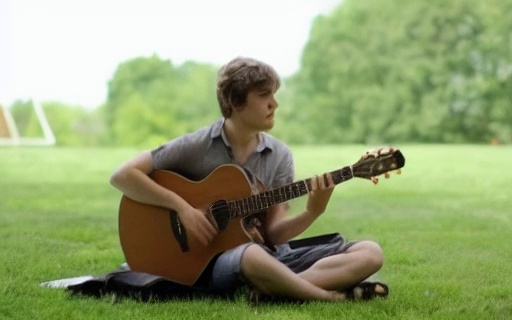} & 
\includegraphics[width=0.10\linewidth]{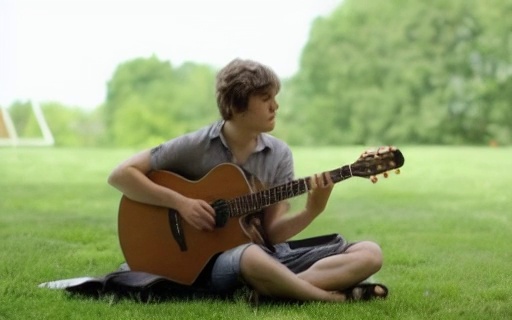}\hspace{2pt} & \\

\raisebox{9pt}{\rotatebox[origin=l]{90}{\small $\caseH$}}\hspace{4pt} &

\includegraphics[width=0.10\linewidth]{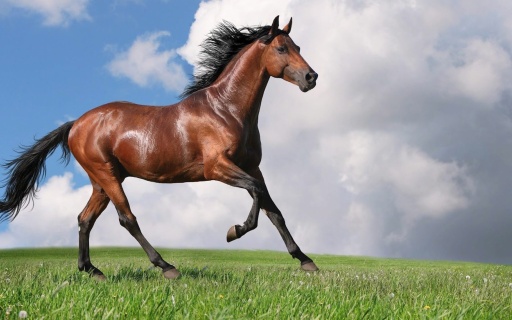}\hspace{2pt} &

\includegraphics[width=0.10\linewidth]{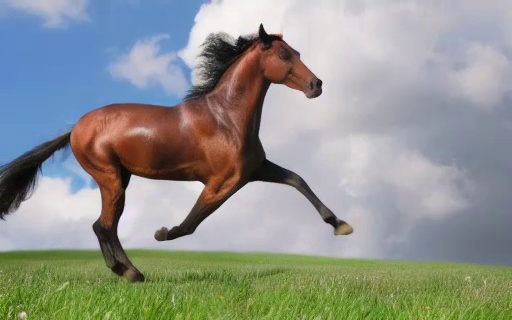} & 
\includegraphics[width=0.10\linewidth]{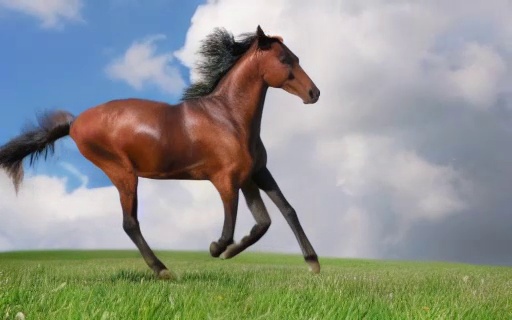}\hspace{2pt} &
\includegraphics[width=0.10\linewidth]{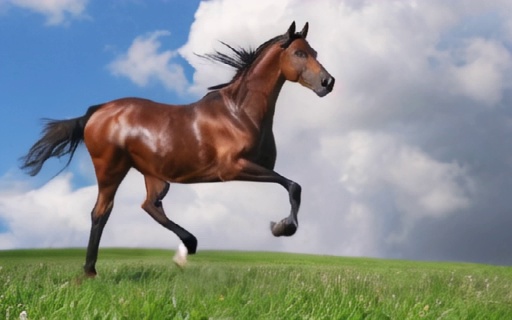} &
\includegraphics[width=0.10\linewidth]{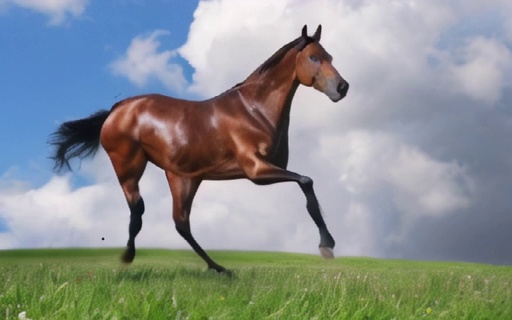}\hspace{2pt} &
\includegraphics[width=0.10\linewidth]{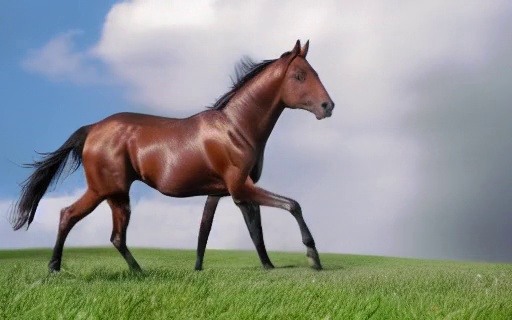} &
\includegraphics[width=0.10\linewidth]{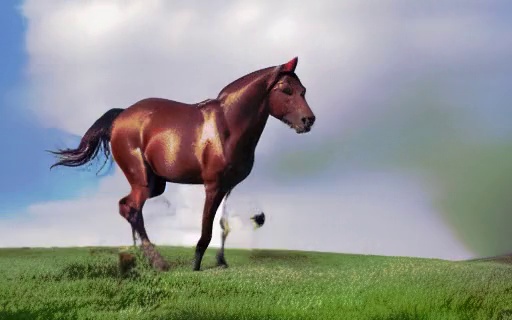}\hspace{2pt} &
\includegraphics[width=0.10\linewidth]{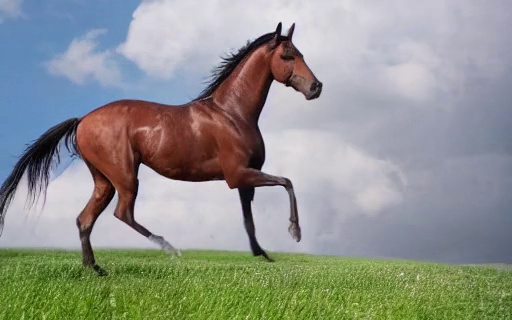} &     
\includegraphics[width=0.10\linewidth]{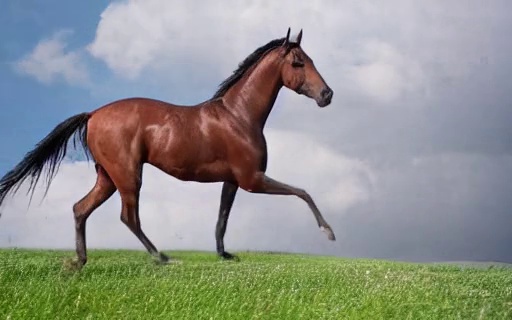}\hspace{2pt} & \\

\raisebox{9pt}{\rotatebox[origin=l]{90}{\small $\caseP$}}\hspace{4pt} &
\includegraphics[width=0.10\linewidth, trim=70 40 58 40, clip]{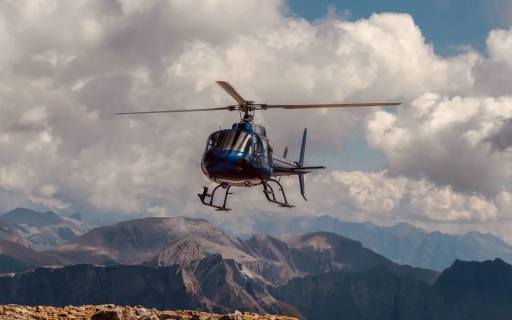}\hspace{2pt} &
\includegraphics[width=0.10\linewidth, trim=70 40 58 40, clip]{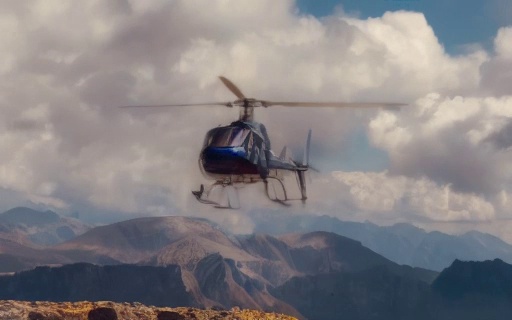} &
\includegraphics[width=0.10\linewidth, trim=70 40 58 40, clip]{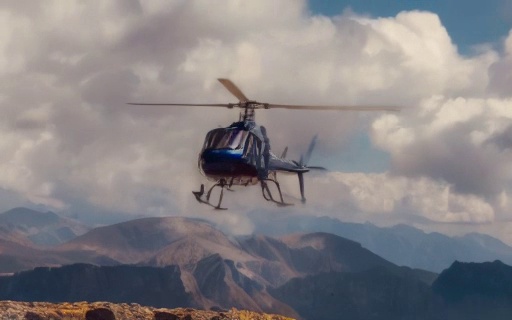}\hspace{2pt} &
\includegraphics[width=0.10\linewidth, trim=70 40 58 40, clip]{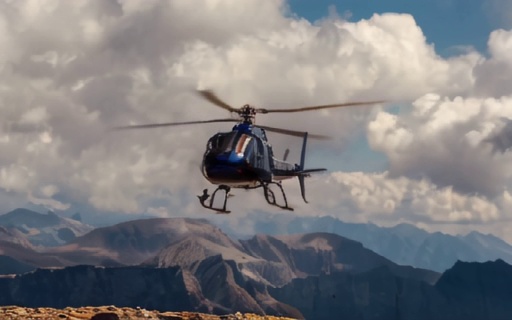} &
\includegraphics[width=0.10\linewidth, trim=70 40 58 40, clip]{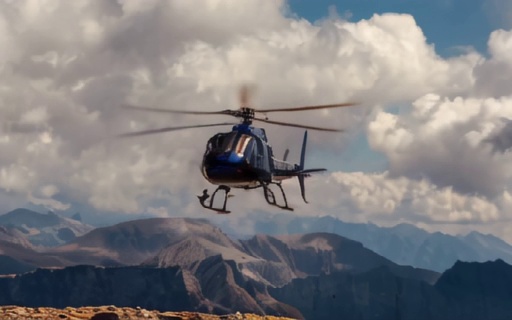}\hspace{2pt} &
\includegraphics[width=0.10\linewidth, trim=70 40 58 40, clip]{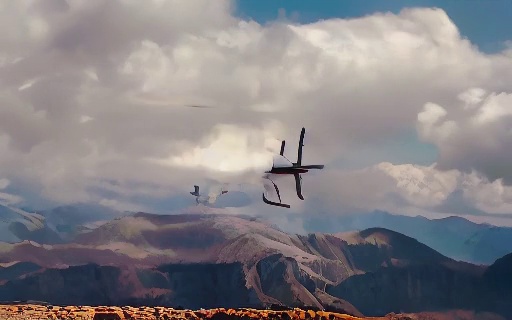} &
\includegraphics[width=0.10\linewidth, trim=70 40 58 40, clip]{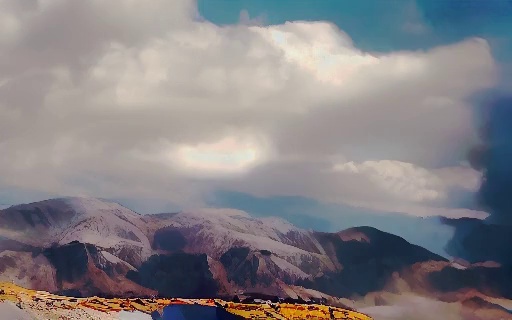}\hspace{2pt} &
\includegraphics[width=0.10\linewidth, trim=70 40 58 40, clip]{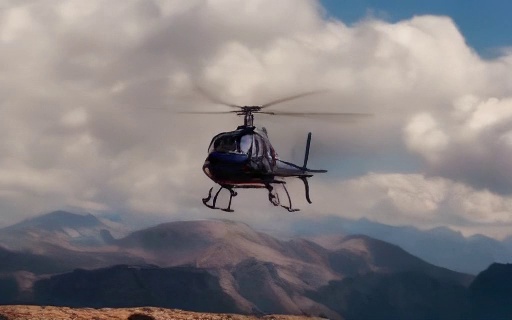} & 
\includegraphics[width=0.10\linewidth, trim=70 40 58 40, clip]{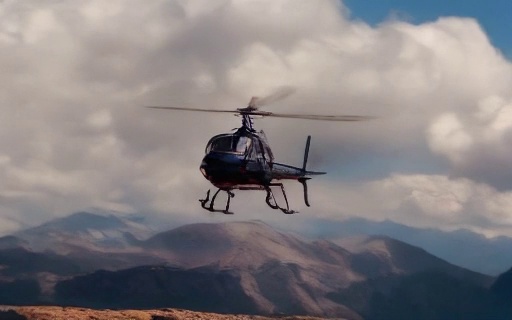}\hspace{2pt} & \\

\raisebox{9pt}{\rotatebox[origin=l]{90}{\small $\caseR$}}\hspace{4pt} &
\includegraphics[width=0.10\linewidth]{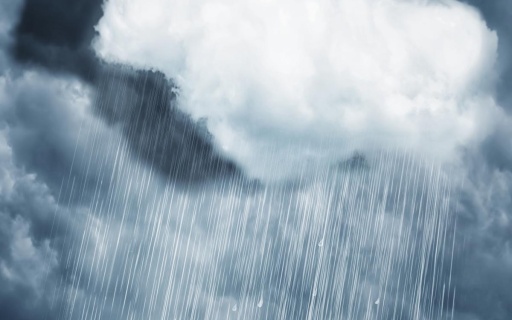}\hspace{2pt} &
\includegraphics[width=0.10\linewidth]{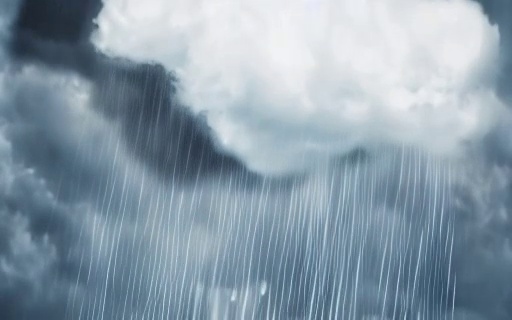} &
\includegraphics[width=0.10\linewidth]{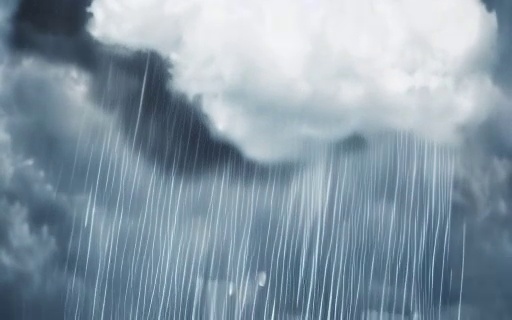}\hspace{2pt} &
\includegraphics[width=0.10\linewidth]{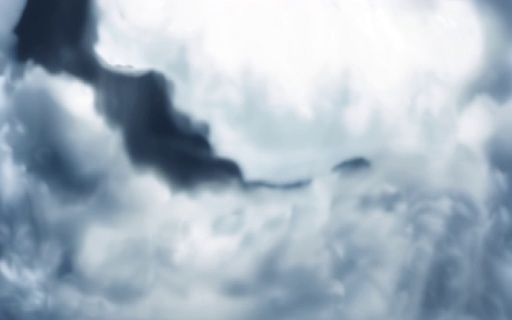} &
\includegraphics[width=0.10\linewidth]{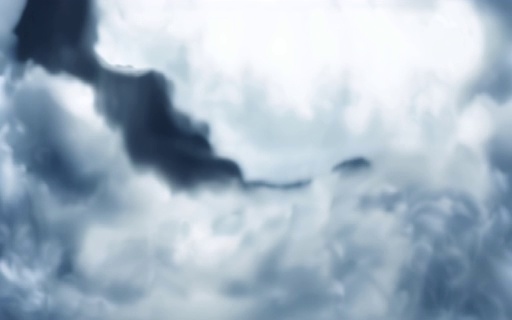}\hspace{2pt} &
\includegraphics[width=0.10\linewidth]{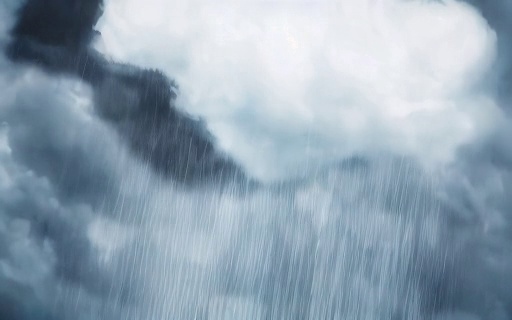} &
\includegraphics[width=0.10\linewidth]{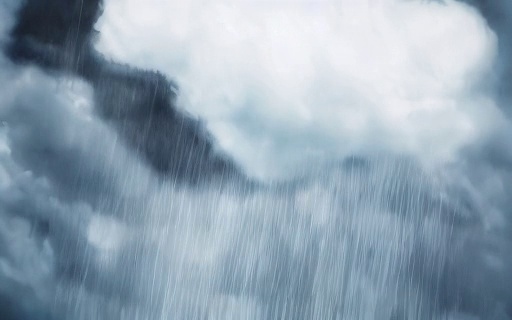}\hspace{2pt} &
\includegraphics[width=0.10\linewidth]{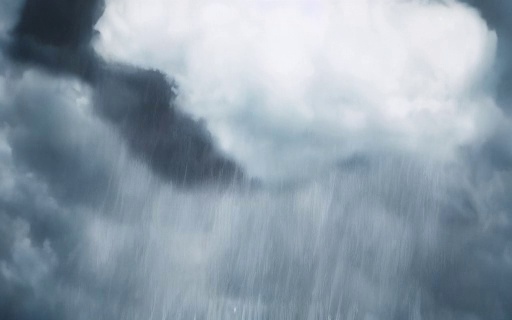} & 
\includegraphics[width=0.10\linewidth]{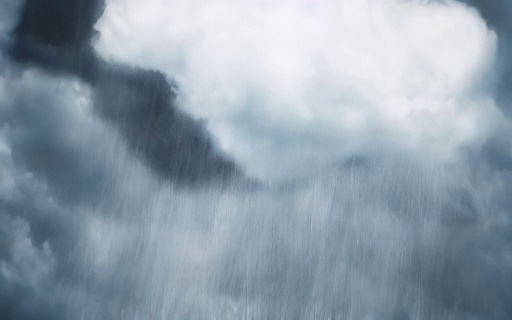}\hspace{2pt} & \\

\raisebox{9pt}{\rotatebox[origin=l]{90}{\small $\caseS$}}\hspace{4pt} &
\includegraphics[width=0.10\linewidth]{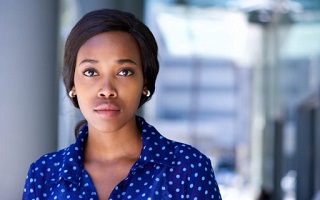}\hspace{2pt} &
\includegraphics[width=0.10\linewidth]{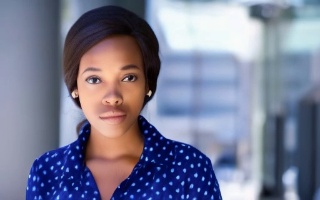} &
\includegraphics[width=0.10\linewidth]{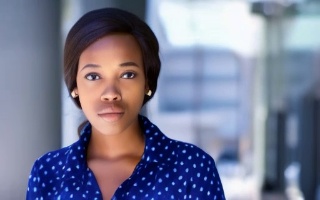}\hspace{2pt} &
\includegraphics[width=0.10\linewidth]{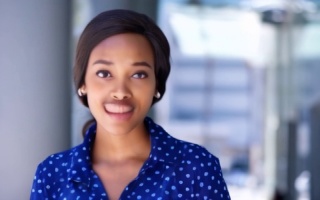} &
\includegraphics[width=0.10\linewidth]{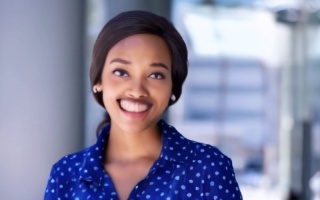}\hspace{2pt} &
\includegraphics[width=0.10\linewidth]{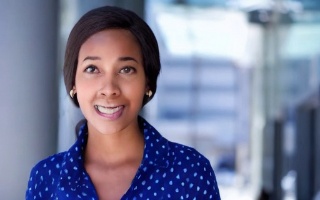} &
\includegraphics[width=0.10\linewidth]{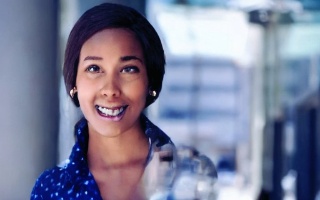}\hspace{2pt} &
\includegraphics[width=0.10\linewidth]{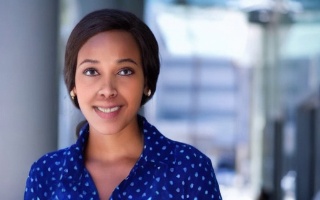} & 
\includegraphics[width=0.10\linewidth]{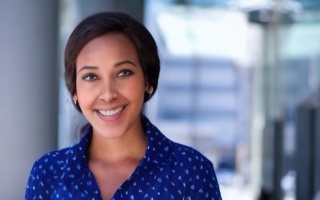}\hspace{2pt} & \\

\raisebox{7pt}{\rotatebox[origin=l]{90}{\small
    $\caseW$}}\hspace{4pt} &

\includegraphics[width=0.10\linewidth, trim=110 50 18 30, clip]{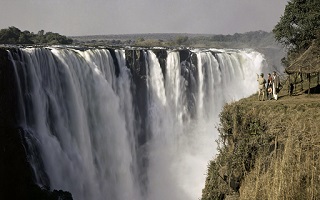}\hspace{2pt} &

\includegraphics[width=0.10\linewidth, trim=110 50 18 30, clip]{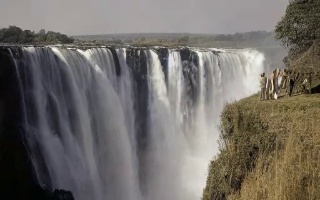} &
\includegraphics[width=0.10\linewidth, trim=110 50 18 30, clip]{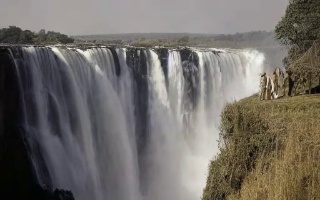}\hspace{2pt} &
\includegraphics[width=0.10\linewidth, trim=110 50 18 30, clip]{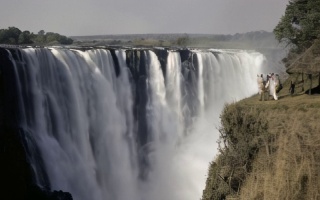} &
\includegraphics[width=0.10\linewidth, trim=110 50 18 30, clip]{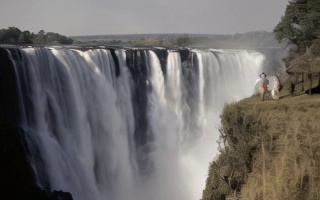}\hspace{2pt} &
\includegraphics[width=0.10\linewidth, trim=110 50 18 30, clip]{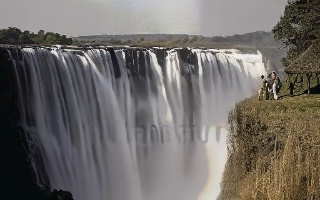} &
\includegraphics[width=0.10\linewidth, trim=110 50 18 30, clip]{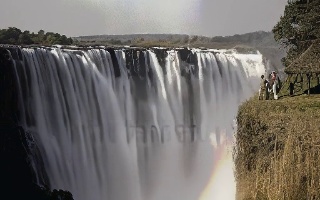}\hspace{2pt} &
\includegraphics[width=0.10\linewidth, trim=110 50 18 30, clip]{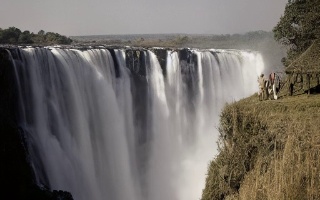} &
\includegraphics[width=0.10\linewidth, trim=110 50 18 30, clip]{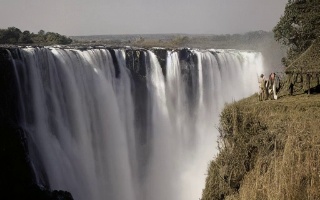}\hspace{2pt} & \\

\end{tabular}

\small \raggedright 
\hspace{14mm} Input \hspace{15mm} I2V-Adapter \hspace{21mm} Cinemo \hspace{25.5mm} LAMP \hspace{26mm} MIVA

\centering
\setlength{\tabcolsep}{0pt} 
\begin{tabular}{c*{10}{c}}
\rotatebox[origin=l]{90}{\small $\caseW$+$\caseH$}\hspace{4pt} &

\includegraphics[width=0.10\linewidth, trim=80 30 48 50, clip]{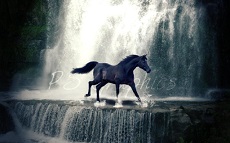}\hspace{2pt} &

\includegraphics[width=0.10\linewidth, trim=80 30 48 50, clip]{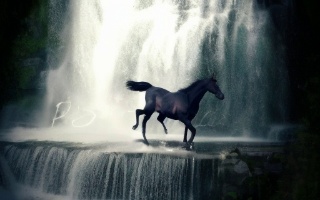} &
\includegraphics[width=0.10\linewidth, trim=80 30 48 50, clip]{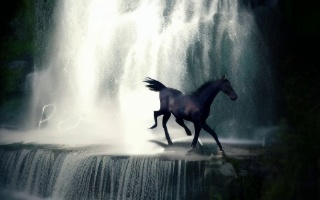}\hspace{2pt} &
\includegraphics[width=0.10\linewidth, trim=80 30 48 50, clip]{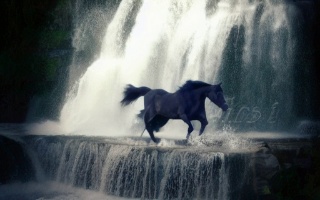} &
\includegraphics[width=0.10\linewidth, trim=80 30 48 50, clip]{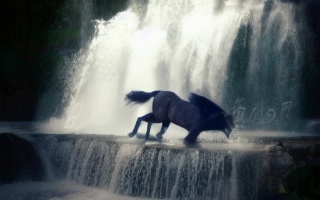}\hspace{2pt} &
\includegraphics[width=0.10\linewidth, trim=80 30 48 50, clip]{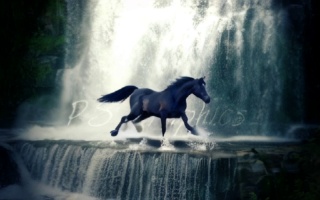} &
\includegraphics[width=0.10\linewidth, trim=80 30 48 50, clip]{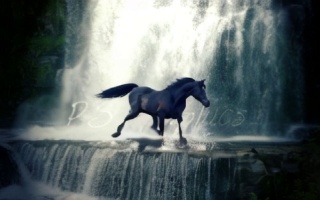}\hspace{2pt} &
\includegraphics[width=0.10\linewidth, trim=80 30 48 50, clip]{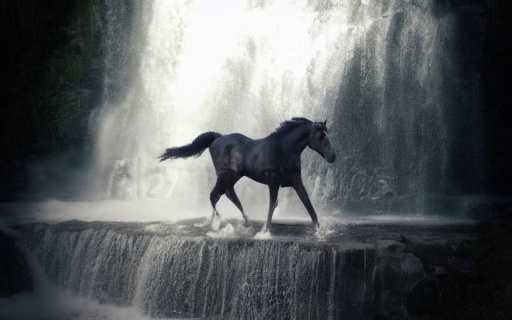} &
\includegraphics[width=0.10\linewidth, trim=80 30 48 50, clip]{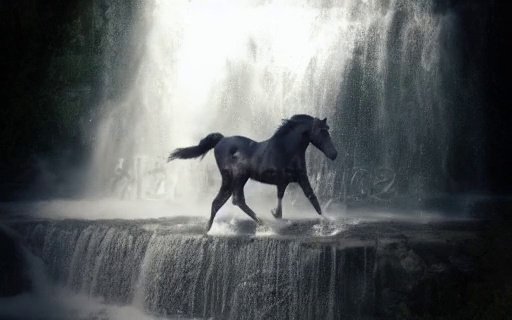}\hspace{2pt} & \\

\rotatebox[origin=l]{90}{\small ~ $\caseG$+$\caseS$}\hspace{4pt} &

\includegraphics[width=0.10\linewidth]{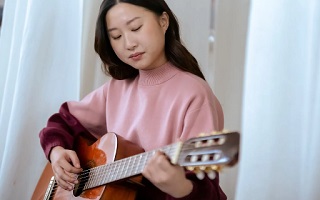}\hspace{2pt} &

\includegraphics[width=0.10\linewidth]{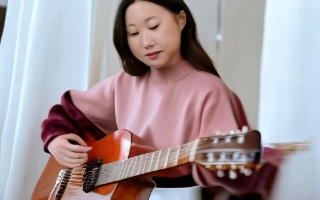} & 
\includegraphics[width=0.10\linewidth]{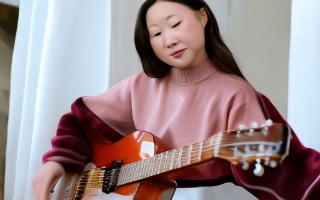}\hspace{2pt} &
\includegraphics[width=0.10\linewidth]{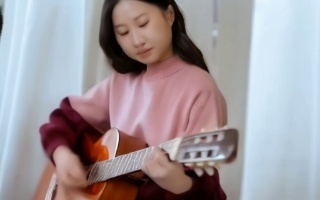} &
\includegraphics[width=0.10\linewidth]{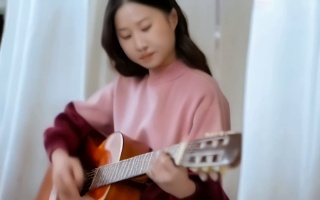}\hspace{2pt} &
\includegraphics[width=0.10\linewidth]{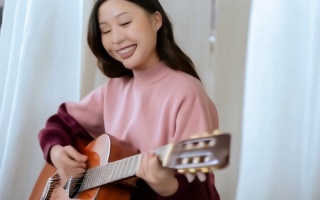} &
\includegraphics[width=0.10\linewidth]{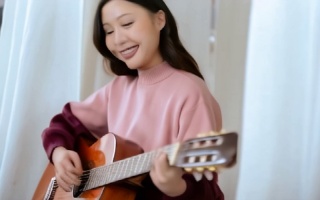}\hspace{2pt} &
\includegraphics[width=0.10\linewidth]{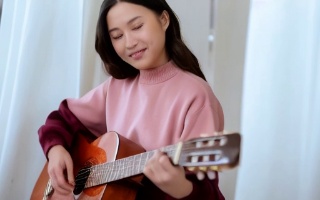} & 
\includegraphics[width=0.10\linewidth]{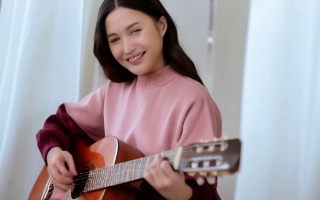}\hspace{2pt} & \\

\end{tabular}

\small \raggedright 
\hspace{14mm} Input \hspace{12mm} DynamicCrafter \hspace{17mm} I2V-Adapter \hspace{21mm} Cinemo \hspace{17mm} MIVA (2 combined)

\caption{Single-motion-pattern (rows 1-8) and multi-motion-pattern (rows 9-10) animation by the I2V DMs.} 
\label{fig:single}

\end{figure*}

\subsubsection{Accelerated Inference}
Due to the comparatively simpler structure of binary masks relative to natural images, we observe that mask sequences converge significantly earlier in the diffusion process than video frames
(see Fig.~\ref{fig:1step}).
Leveraging this, we bypass mask generation during a large portion of time steps by reusing the buffered attention masks computed earlier, thereby greatly reducing computational overhead without degrading the predicted masks $\tilde S$.

\subsection{Parallelizing Multiple MIVAs}
\label{s:parallel}

In multi-motion-pattern animation cases, MIVAs can be plugged into the base DM in parallel to collaborate,
without requiring test-time fine-tuning.
This is facilitated by the residual learning mechanism \cite{resnet} 
featured by the skip connections in DMs, guiding each layer to predict the residual of the target signal rather than the signal itself. 
Consequently, residual outputs from different MIVAs can be summed without domain deviation. 
Moreover, the summation can incorporate individual weights for each MIVA to modulate the intensity of the associated motion.
More details can be found in Supplementary Materials (SM).

\section{Experiments}
\label{s:exp}

\subsection{Experimental Setup}
\label{s:exp_setup}

\subsubsection{Data}
To ensure a fair comparison with LAMP, we strictly adhere to the training dataset curated by the authors.
The dataset consists of 8 motion patterns: birds flying $\caseB$, fireworks $\caseF$, guitar playing $\caseG$, horse running $\caseH$, helicopter $\caseP$, raining $\caseR$, turning to smile $\caseS$ and waterfall $\caseW$, with 8-16 short videos for each.
For each training iteration, we randomly sample a 16-frame clip (equivalent to 2 seconds) from the selected video
and resize the clip to the resolution of $512\times 320$. 

For testing, we construct benchmark datasets for single-motion-pattern and multi-motion-pattern settings respectively.
Different from \cite{lamp} that evaluates LAMP by animating 6 DM-generated images per category, our benchmark datasets are fully formed by real-world images.
The single-motion-pattern benchmark dataset includes 20 images per category, each featuring a motion subject with moderate diversity. The multi-motion-pattern benchmark dataset consists of 6 dual-motion-pattern tasks, with 5-10 images per task due to limited data availability.


\subsubsection{Implementation}
We implement MIVA based on the T2V DM AnimateDiff-v3 \cite{animatediff}. All experiments are conducted using a single NVidia V100 GPU, with a VRAM usage as low as 9.8GB for training.
The learnable parameters constitute 3\% of the base DM, increasing to 5\% for M-MIVA.
More details are provided in SM.





\subsection{Comparison Study}
We compare our MIVA-powered I2V models (without masks) with LAMP along with several recent I2V DMs, all based on the same LDM framework, in chronological order:
Stable Video Diffusion (SVD) \cite{svd}, 
I2VGen-XL \cite{i2vgen}, 
DynamiCrafter \cite{dynamicrafter}, 
I2V-Adapter \cite{i2v-adapter}, 
and Cinemo \cite{cinemo}. 
The methods except LAMP are all trained on WebVid-10M
\cite{webvid10m} with 10M samples
, aiming at open domain image animation.
Cinemo is further trained on 25M additional samples curated by its authors.
When deploying each method, we follow the respective default setting, without manual hyperparameter tuning.
Since most methods rely on text prompts, we prepare custom prompts for each input image with both the target subject(s) and the desired motion.


We assess model performance using VBench \cite{vbench}, which provides both temporal and frame-wise quality metrics. Temporal metrics include background consistency, motion smoothness, subject consistency, and temporal flickering; frame-wise metrics cover aesthetic quality and image quality. To further evaluate motion intensity, we utilize EvalCrafter \cite{evalcrafter} to compute the average flow score. Additionally, we perform a user study for subjective evaluation, yielding a preference rate for each method.
More setup details are presented in SM.


\subsubsection{Single-motion-pattern Animation}
\label{s:single}

Tab.~\ref{tab:main} presents the evaluation result on our single-motion-pattern benchmark dataset. With the same training data, our method significantly outperforms LAMP and even matches the performance of state-of-the-art I2V models, despite requiring substantially fewer training samples.
The user study shows a significant advantage of our method over others.

Fig.~\ref{fig:single} illustrates representative animation results, with additional examples provided in SM.
Viewers consistently reported three types of failure modes with baseline I2V methods.
First, not following the prompt: 
models such as SVD, I2VGen-XL, DynamiCrafter and I2V-Adapter frequently introduce uncontrollable camera motion, while Cinemo tends to generate insufficient motion (\eg, case $\caseR$), as evidenced by its extremely low average flow score.
Second, object hallucination: several methods exhibit abrupt object alterations or deletions, compromising visual integrity. For example, most models incorrectly modify human subjects in cases $\caseF$ and $\caseW$; LAMP creates a watermark to the waterfall in $\caseW$, and an irrelevant white box in $\caseG$.
Third, motion unrealism: generated motion often diverges from physical sense, resulting in unnatural or exaggerated behavior, such as the horse legs in $\caseH$ and smiling faces in $\caseS$ that are severely distorted.
In contrast, our method demonstrates strong resilience to these issues, consistently producing physically plausible animations that align with the user’s intended motion.


\subsubsection{Multi-motion-pattern Animation}
\label{s:multi}

The bottom two rows of Fig.~\ref{fig:single} are cases from our multi-motion-pattern benchmark set, each being a combination of two motion patterns out of the four in Sec.~\ref{s:single}.
We parallelize two MIVAs for each case with equal weights of $0.5$, and exclude LAMP due to its incapability.
Aside from the issues in single-motion-pattern scenarios, 
methods trained on large datasets often struggle to follow text prompts with multiple motion patterns,
resulting in uncontrollable dynamics, sometimes generating only one motion pattern or none at all.
With a divide-and-conquer strategy, integrating two MIVAs effectively resolves the challenges of multi-pattern animation, offering enhanced control and reliable synthesis.
Furthermore, even when scaling to multiple parallel MIVAs,
our approach maintains consistent quantitative performance and continues to outperform the existing methods in user preference evaluations.






\subsection{Ablation Study}


\subsubsection{CFA-Weighting Modules $\phi$}

\begin{figure}[t]
  \centering
  \includegraphics[width=0.85\linewidth]{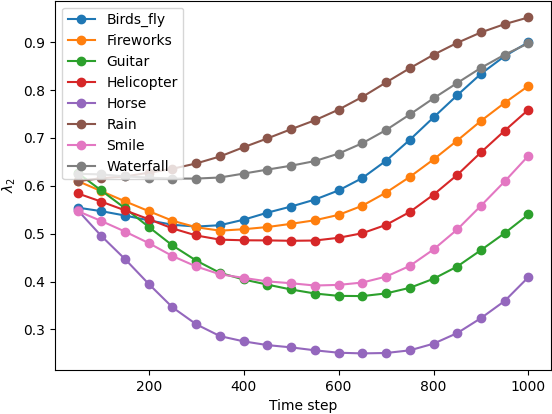}
   \caption{Learned $CFA_{i,1}$ weight, $\lambda_2$, of the U-Net's bottommost attention block across different motion patterns.
   }
   \label{fig:sa_weight}
\end{figure}

We observe substantial variation in the CFA weights $\{\lambda_{2,3}\}$ across layers, time steps, and motion types.
We select the bottleneck block of the U-Net, of lowest resolution and thus conveying most semantic information, to shed a light on the association between $\lambda_2$ and the motion pattern. 
A larger $\lambda_2$ indicates stronger dependency on the input image $I$ and, on the other hand, less dependency on the previous frame. 
As shown in Fig.~\ref{fig:sa_weight},
the rain and waterfall cases exhibit the highest $\lambda_2$, indicating minimal reliance on the previous frame due to the complex, fast-evolving textures of fluid dynamics.
Conversely, $\lambda_2$ are lowest for horse and human motion, consistent with the expectation that body and facial animation require greater temporal coherency, wherein the previous frame plays a more important role.

\begin{figure}[t]
\centering
\setlength{\tabcolsep}{0pt} 
\begin{tabular}{c*{6}{c}}

 \raisebox{4pt}{\rotatebox[origin=l]{90}{\small
$\caseR$}}\hspace{4pt} &

\includegraphics[width=0.18\linewidth]{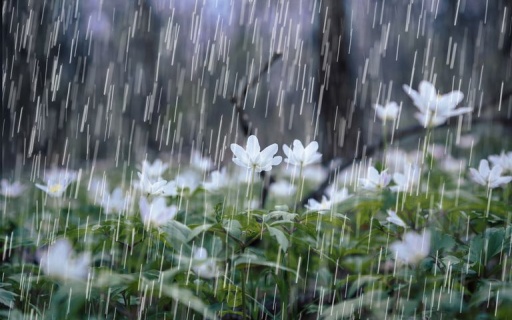}\hspace{2pt} &

\includegraphics[width=0.18\linewidth]{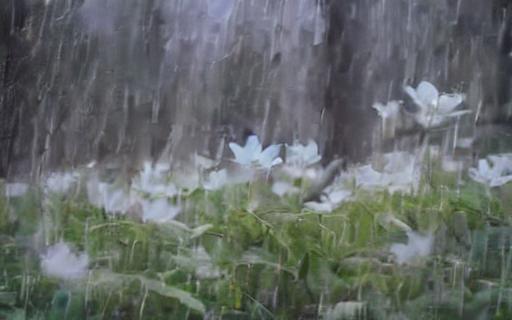} &
\includegraphics[width=0.18\linewidth]{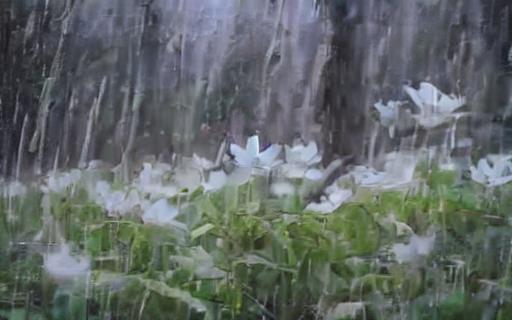} \hspace{2pt} &
\includegraphics[width=0.18\linewidth]{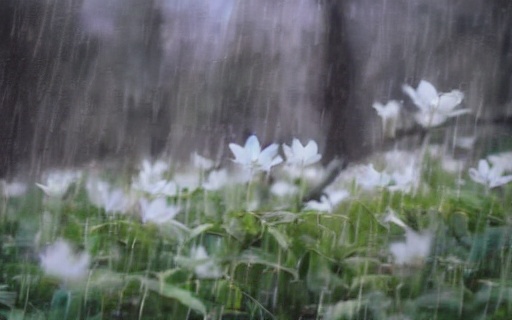} &
\includegraphics[width=0.18\linewidth]{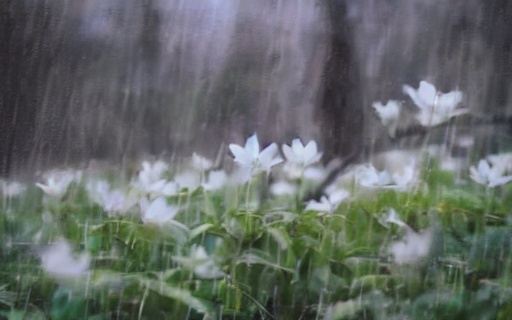} \hspace{2pt} & \\

\raisebox{5pt}{\rotatebox[origin=l]{90}{\small
$\caseH$}}\hspace{4pt} &

\includegraphics[width=0.18\linewidth]{figs/single/horse6/6.jpg}\hspace{2pt} &

\includegraphics[width=0.18\linewidth]{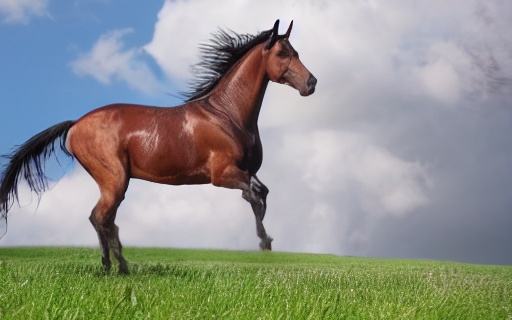} &
\includegraphics[width=0.18\linewidth]{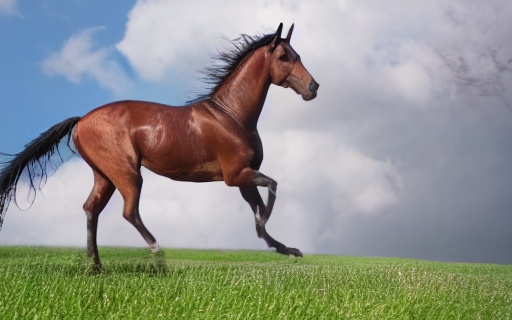} \hspace{2pt} &
\includegraphics[width=0.18\linewidth]{figs/single/horse6/ours_f1.jpg} &
\includegraphics[width=0.18\linewidth]{figs/single/horse6/ours_f2.jpg} \hspace{2pt} & \\
\end{tabular}
\parbox{\linewidth}{
        \hspace{0.09\linewidth}
        \small{Input}
        \hspace{0.14\linewidth}
        \small{Without $\phi$}
        \hspace{0.22\linewidth}
        \small{With $\phi$}
    }\\
\caption{Ablation study on CFA weighting modules $\phi$.}
\label{fig:ablation_sa}

\end{figure}

As shown in Fig.~\ref{fig:ablation_sa}, we conduct a comparison study to assess the impact of $\phi$, where MIVAs with 
$\lambda_2 \equiv \lambda_3 \equiv 0.5$ serve as the baseline.
In case $\caseR$, the adaptive mechanism assigns larger $\lambda_2$, emphasizing the input frame $I$ and promoting better preservation of visual details, while the baseline disproportionately attends to raindrops, resulting in background blur.
Meanwhile, case $\caseH$ poses a greater challenge due to subtle limb movements; existing approaches and the baseline often produce anatomically inconsistent outputs with unrealistic leg counts or body structure.
By biasing the CFA weights toward the previous frame, the adaptive mechanism enhances motion continuity and significantly improves anatomical fidelity 
in maintaining the realism of body appearance, especially that of the legs.
Quantitatively, the introduction of $\phi$ yields consistent gains across most objective evaluation metrics, as reported in Tab.~\ref{tab:main}.


\begin{table}[t]
    \centering
    \footnotesize
    \begin{tabular}{l|ccc}
        \toprule
        \makecell{Method} & \makecell{Subject \\ Consistency $\uparrow$} & \makecell{Aesthetic \\ Quality Loss $\downarrow$} & \makecell{User \\ Preference (\%) $\uparrow$} \\
        \midrule

        LAMP
        & $89.4$ 
        & $5.11$ 
        & $N/A$ \\

        MIVA
        & $94.3$ 
        & $4.47$ 
        & $41.1$ \\

        M-MIVA
        & $95.3$ 
        & $2.93$ 
        & $58.9$ \\
        
        \bottomrule
    \end{tabular}
    \caption{Quantitative comparison of MIVA and M-MIVA on the 3 selected motions ($\caseF$, $\caseB$, $\caseH$).} 
    \label{tab:ablation_mask}
\end{table}

\begin{figure}[t]
    \centering
    \raisebox{3pt}{\rotatebox[origin=l]{90}{\small
    $\caseF$}}\hspace{4pt}
    \includegraphics[width=0.13\linewidth]{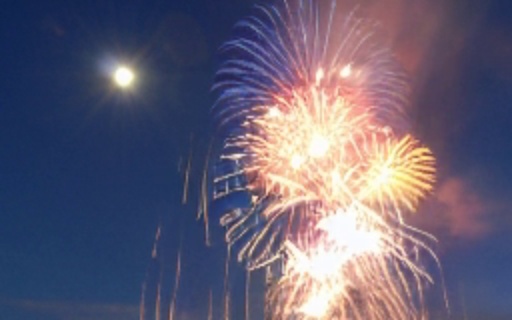}
    \includegraphics[width=0.13\linewidth]{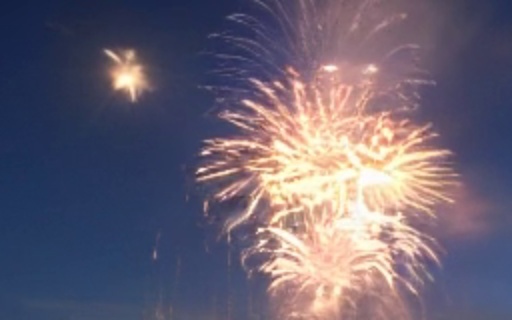}%
    \includegraphics[width=0.13\linewidth]{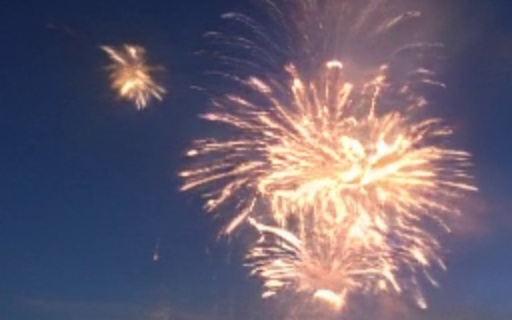}
    \includegraphics[width=0.13\linewidth]{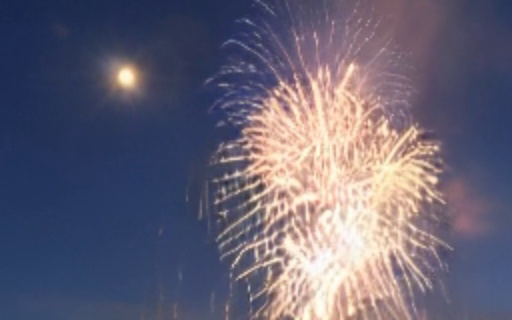}%
    \includegraphics[width=0.13\linewidth]{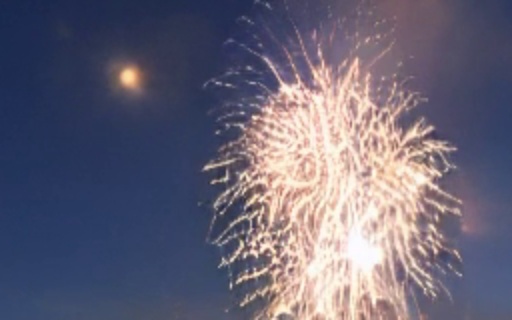}
    \includegraphics[width=0.13\linewidth]{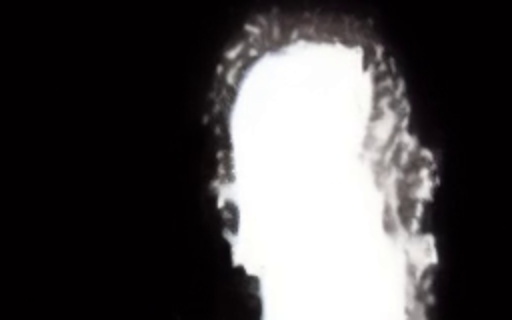}%
    \includegraphics[width=0.13\linewidth]{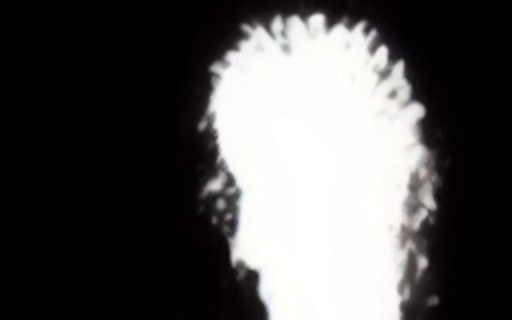}\\
    \raisebox{3pt}{\rotatebox[origin=l]{90}{\small
    $\caseB$}}\hspace{4pt}
    \includegraphics[width=0.13\linewidth]{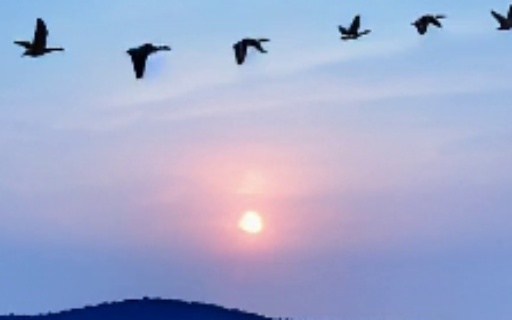}
    \includegraphics[width=0.13\linewidth]{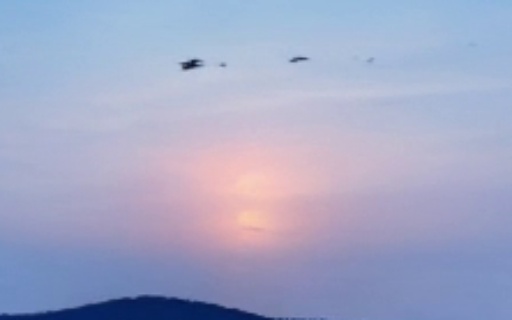}%
    \includegraphics[width=0.13\linewidth]{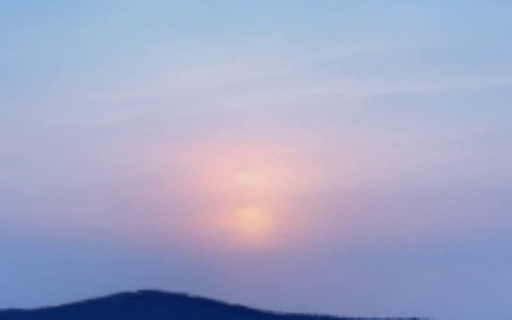}
    \includegraphics[width=0.13\linewidth]{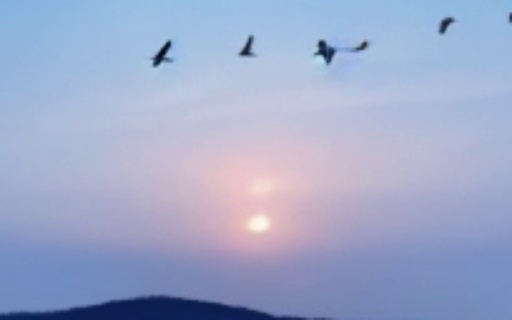}%
    \includegraphics[width=0.13\linewidth]{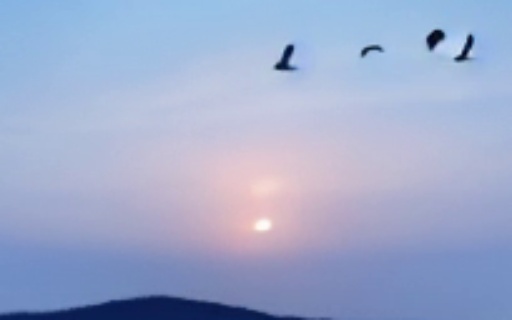}
    \includegraphics[width=0.13\linewidth]{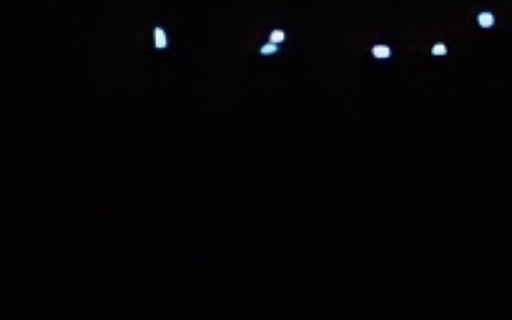}%
    \includegraphics[width=0.13\linewidth]{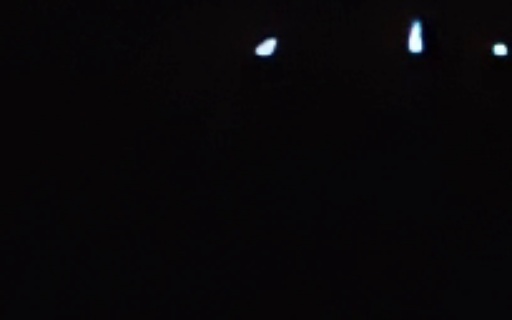}\\
    \raisebox{3pt}{\rotatebox[origin=l]{90}{\small
    $\caseH$}}\hspace{4pt}
    \includegraphics[width=0.13\linewidth]{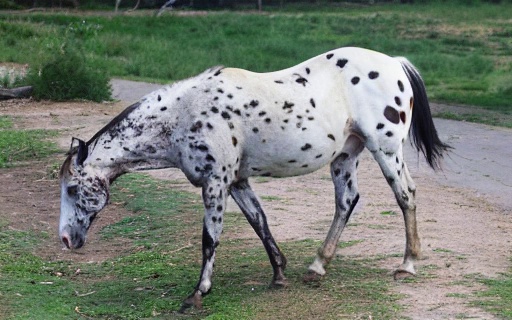}
    \includegraphics[width=0.13\linewidth]{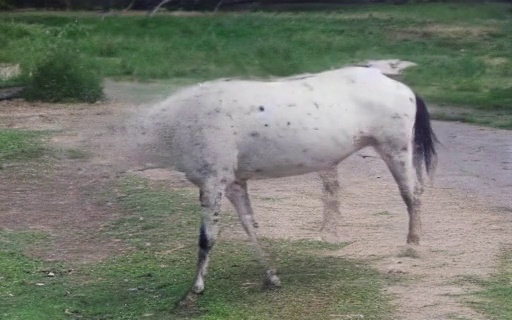}%
    \includegraphics[width=0.13\linewidth]{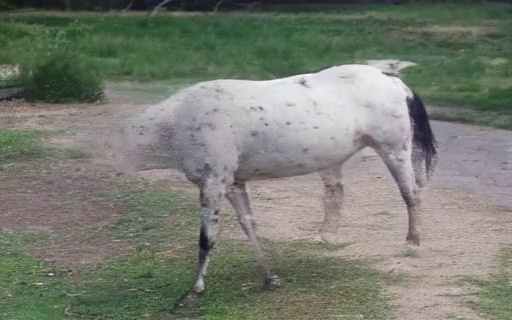}
    \includegraphics[width=0.13\linewidth]{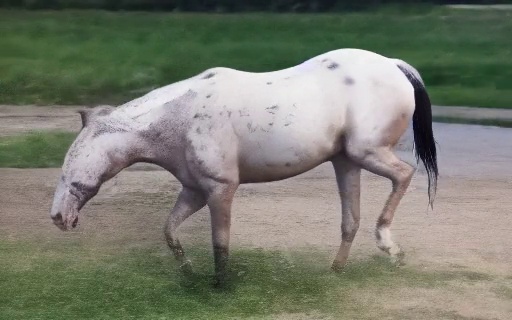}%
    \includegraphics[width=0.13\linewidth]{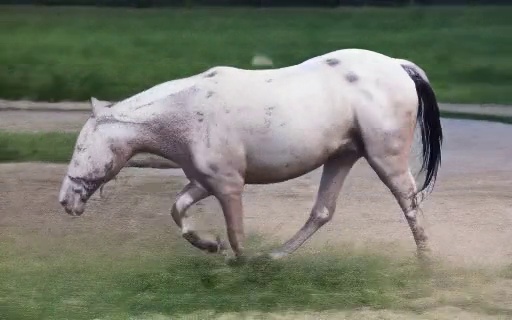}
    \includegraphics[width=0.13\linewidth]{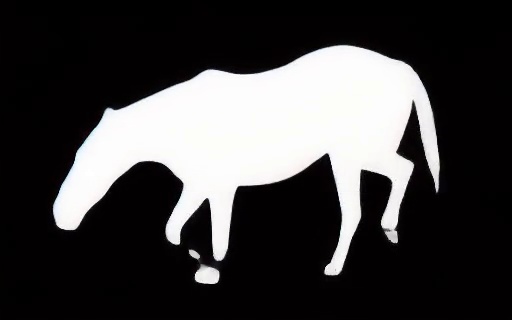}%
    \includegraphics[width=0.13\linewidth]{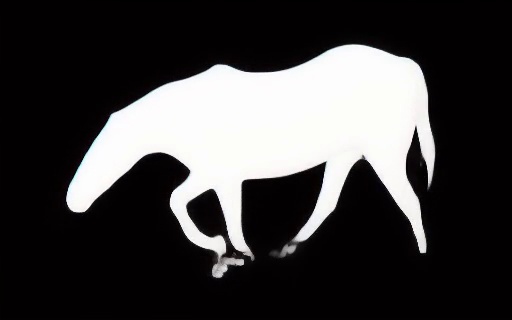}\\
    \parbox{\linewidth}{
        \hspace{0.07\linewidth}
        \small{Input}
        \hspace{0.1\linewidth}
        \small{MIVA}
        \hspace{0.18\linewidth}
        \small{M-MIVA ($V$ and $S$)}
    }\\
    \caption{Ablation study on MIVA versus M-MIVA.}
    \label{fig:ablation_mask}
\end{figure}

\begin{figure}[!t]
    \centering
    \parbox{0.49\linewidth}{
        \centering
        \includegraphics[width=0.32\linewidth]{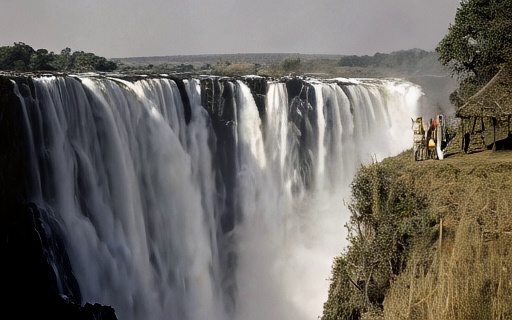}
        \includegraphics[width=0.32\linewidth]{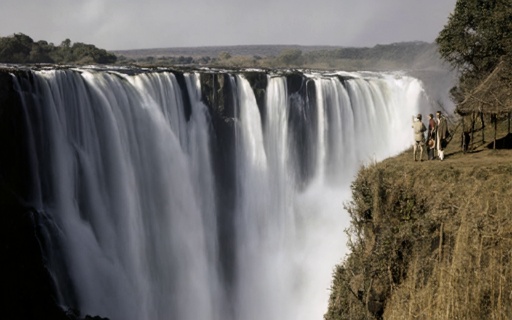}%
        \includegraphics[width=0.32\linewidth]{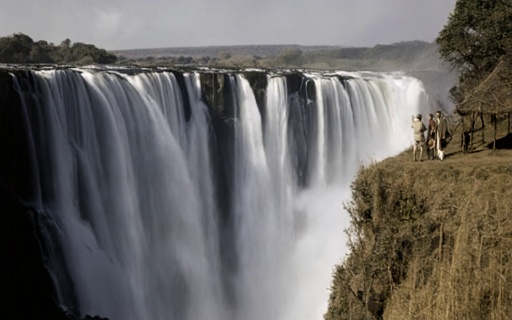}%
    }
    \parbox{0.49\linewidth}{
        \centering
        \includegraphics[width=0.32\linewidth]{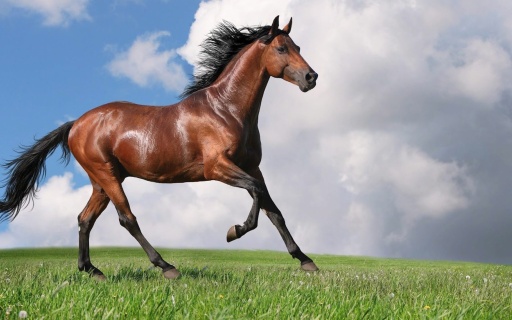}
        \includegraphics[width=0.32\linewidth]{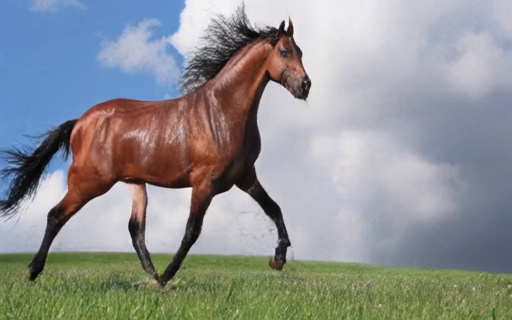}%
        \includegraphics[width=0.32\linewidth]{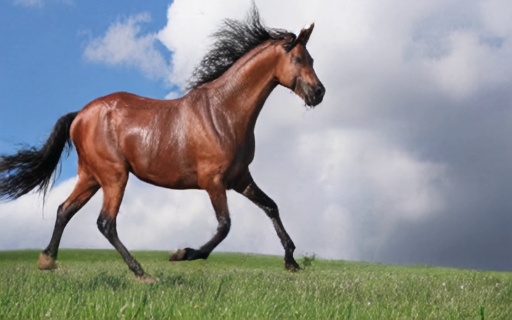}%
    }\\
    \parbox{0.49\linewidth}{
        \centering
        \includegraphics[width=0.32\linewidth]{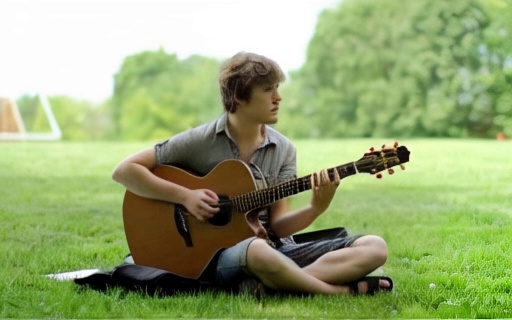}
        \includegraphics[width=0.32\linewidth]{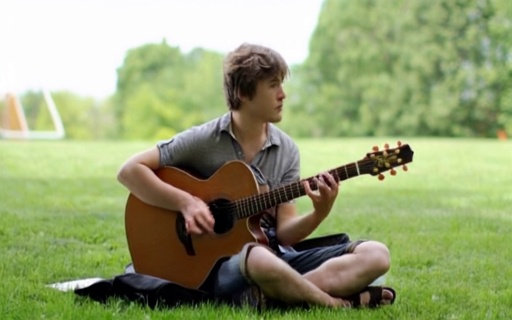}%
        \includegraphics[width=0.32\linewidth]{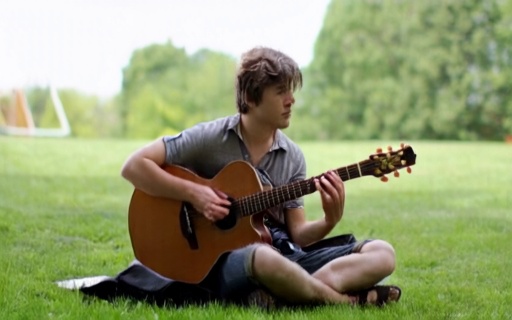}%
    }
    \parbox{0.49\linewidth}{
        \centering
        \includegraphics[width=0.32\linewidth]{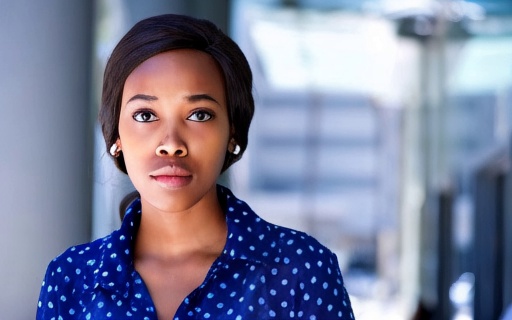}
        \includegraphics[width=0.32\linewidth]{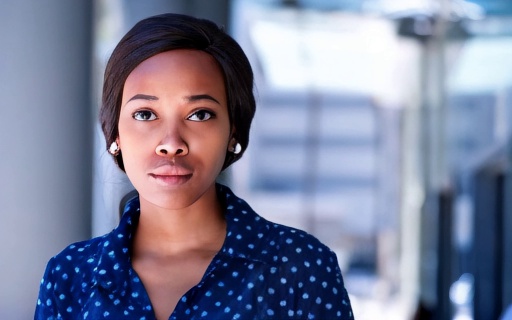}%
        \includegraphics[width=0.32\linewidth]{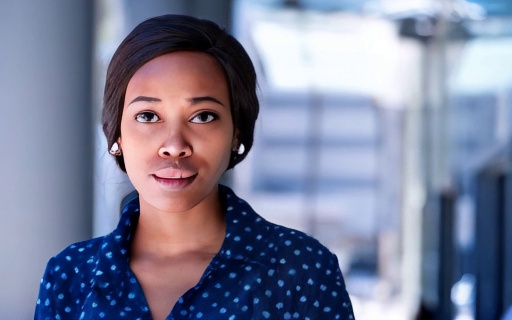}%
    }\\
    \parbox{\linewidth}{
        \hspace{0.04\linewidth}
        \small{Input}
        \hspace{0.14\linewidth}
        \small{MIVA}
        \hspace{0.14\linewidth}
        \small{Input}
        \hspace{0.14\linewidth}
        \small{MIVA}
    }\\
    \caption{Results of MIVA based on Wan2.1-T2V-1.3B.}
    \label{fig:wan}
\end{figure}

\subsubsection{M-MIVA}

Fig.~\ref{fig:ablation_mask} compares vanilla MIVA and M-MIVA on selected motion types: $\caseF$, $\caseB$, and $\caseH$.
In $\caseF$ and $\caseB$, MIVA exhibits motion leakage:
the moon is erroneously animated as fireworks, and both the birds and the sun fade due to interference from the background sky.
In contrast, M-MIVA effectively mitigates these issues by leveraging the generated subject mask sequence to confine motion synthesis to relevant regions.
Case $\caseH$ reveals the impact of dataset bias:
LAMP training samples for $\caseH$ predominantly feature dark horses, causing MIVA to overfit and produce artifacts when animating a white horse. In this example, parts of the subject are misclassified as background, resulting in illogical animation.
Subject mask sequences exhibit greater robustness to such biases by clearly delineating the motion subject from its surroundings, substantially improving generalization.
Quantitative evaluation and a head-to-head user study (Tab.~\ref{tab:ablation_mask}) further validate the contribution of mask guidance to temporal coherence and visual fidelity.
Nonetheless, M-MIVA shows limited improvement on the five remaining motions
in the LAMP dataset. We provide detailed discussions in SM.

\subsection{MIVA for DiT}
To demonstrate the compatibility with DiT, we implement MIVAs for another DiT-based T2V model, Wan-T2V-1.3B \cite{wan}. 
Results are shown in Fig.~\ref{fig:wan}. More details and discussions are presented in SM.

\section{Conclusion}

We study the paradigm of modular I2V, 
handling a motion pattern via a dedicated MIVA, a lightweight 
sub-network attachable to pre-trained T2V DMs. 
MIVA enables efficient training with minimal data 
and benefits from joint mask sequence generation.
Its modular design grants desirable motion controllability and fusion.
Experiments show promising performance, rivaling models trained on massive data.
We demonstrate the ease of deployment and parallelism of MIVA,
exhibiting comparable or even superior generation performance versus state-of-the-art methods while significantly reducing training and data costs.

{
    \small
    \bibliographystyle{ieeenat_fullname}
    \bibliography{refs}
}

\clearpage
\newpage
\setcounter{section}{0}
\renewcommand{\theequation}{S\arabic{equation}}
\renewcommand{\thefigure}{S\arabic{figure}}
\renewcommand{\thetable}{S\arabic{table}}
\renewcommand{\thesection}{\Alph{section}}

\section{Attachments}

We invite readers to view the image animation results in video format. 
Please refer to the project page (referred to as ``Project Page'' below) at: \url{https://yishaohan.github.io/MIVA-web}.

The codes are released at: \url{https://github.com/yishaohan/MIVA}.


\section{Additional Details of MIVA}

\subsection{Masked MIVA (M-MIVA)}

\subsubsection{Resolution Alignment for Attention Masks}

The decoder from the pre-trained variational autoencoder (VAE) typically expands the feature map resolution by a factor of four.
To ensure compatibility with the dimensions of $QK^T$ at each attention layer, 
we reshape the attention mask accordingly using bilinear interpolation.


\subsubsection{Scheduling of $p$ in Dropout Training}

When training M-MIVA, with probability $p$, we use the ground truth subject mask sequence $\{S_0^{1:F}\}$ in place of the one-step prediction $\tilde S$.
$p$ follows a cosine decay schedule with respect to the training iteration step $t_{\text{train}}$: $p(t_{\text{train}}) = \frac{1}{2} \left(1 + \cos\left(\frac{t_{\text{train}}}{t_{\text{max}}} \pi \right)\right)$,
where $t_{\text{max}}$ is the total number of training iterations.



\subsubsection{Accelerated Inference}

To mitigate the additional computational burden introduced by the mask modality, we strategically reduce the frequency of mask generation.
Specifically, during the 50-step DDIM sampling process \cite{ddim}, subject masks are generated only at time steps $\{0, 5, \dots, 35 \}$,
totaling 8 computations.
For each of these steps, attention masks are derived and cached.
At the remaining time steps, the previously cached attention masks are reused to modulate the attention layers.

\begin{figure}[t]
    \centering
    \raisebox{-1pt}{\rotatebox[origin=l]{90}{\small
    50 steps}}\hspace{4pt}
    \includegraphics[width=0.22\linewidth]{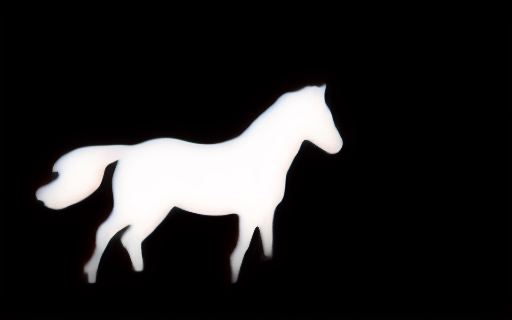}
    \includegraphics[width=0.22\linewidth]{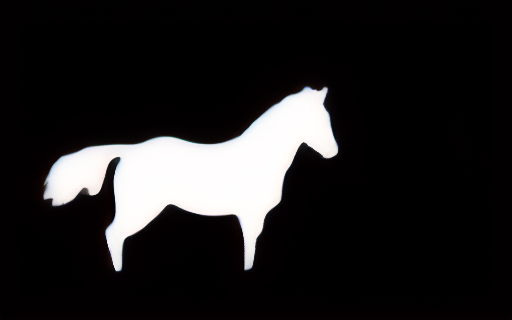}
    \includegraphics[width=0.22\linewidth]{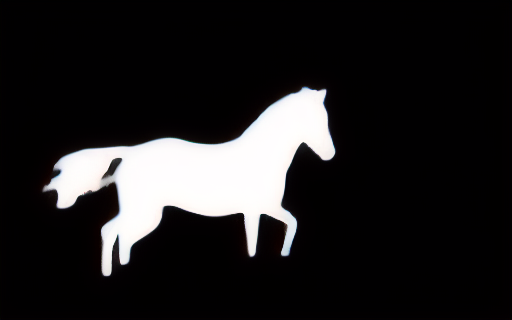}
    \includegraphics[width=0.22\linewidth]{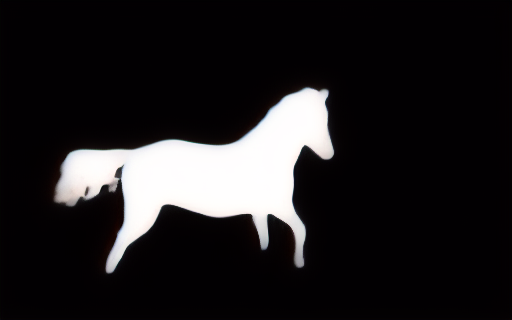}\\
    \raisebox{1pt}{\rotatebox[origin=l]{90}{\small
    8 steps}}\hspace{4pt}
    \includegraphics[width=0.22\linewidth]{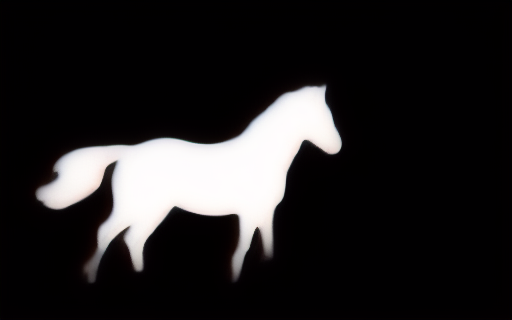}
    \includegraphics[width=0.22\linewidth]{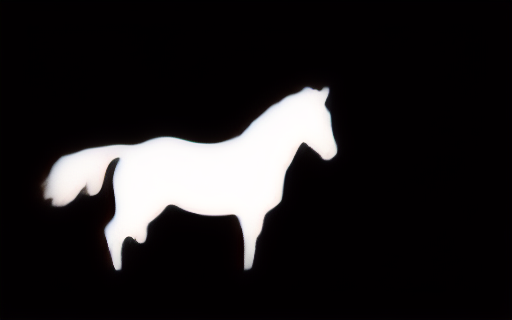}
    \includegraphics[width=0.22\linewidth]{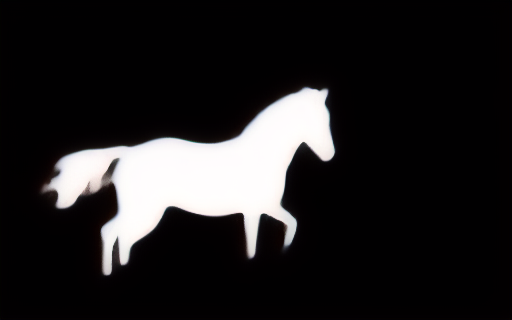}
    \includegraphics[width=0.22\linewidth]{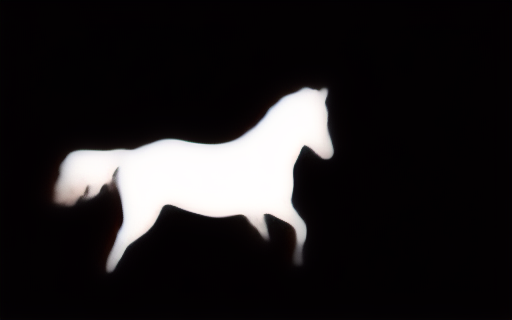}\\
    \parbox{\linewidth}{
        \hspace{0.11\linewidth}
        \small{Frame 4}
        \hspace{0.08\linewidth}
        \small{Frame 8}
        \hspace{0.09\linewidth}
        \small{Frame 12}
        \hspace{0.06\linewidth}
        \small{Frame 16}
    }\\
    \caption{Comparison between 50-step and 8-step mask generation.}
    \label{fig:supp_mask}
\end{figure}

As shown in Fig.~\ref{fig:supp_mask}, this sparse 8-step mask generation yields results nearly indistinguishable from those generated by 50 steps.
While a minor blur is observed in the masks,
we find no discernible difference in the generated frames.

We summarize the average inference time over 20 runs on a single V100 GPU under various MIVA configurations:
\begin{itemize}
    \item Base DM (AnimateDiff-v3), for text-to-video generation: 40.04s;
    \item Base DM + MIVA: 44.50s;
    \item Base DM + M-MIVA with 50-step mask generation: 114.25s;
    \item Base DM + M-MIVA with 8-step mask generation: 54.53s.
\end{itemize}


\subsection{Parallelizing Multiple MIVAs}
\label{s:supp_parallel}

We detail the computation procedure for parallelizing multiple MIVAs.
For cross-attention (CA) and temporal self-attention (t-SA) layers,
the outputs from individual MIVAs are combined via a weighted summation.
In contrast,
spatial self-attention (SA) layers in U-Nets and SA layers in DiTs require special handling, 
as all MIVAs share the same pre-trained SA layer at each block.
Given a base SA layer and $n$ MIVAs, each associated with a weight $w_j, j=1, \dots, n$,
the output for the $i$-th frame from the CFA-infused SA layer is computed as:
\begin{equation}
\lambda_1 SA(\bff^i) + \sum_{j=1}^n  w_j \left[ \lambda_2^{(j)} CFA^{(j)}_{i, 1} + \lambda_3^{(j)} CFA^{(j)}_{i, i-1} \right],
\label{eq:supp}
\end{equation}
where the superscript $(j)$ identifies contributions from the $j$-th MIVA instance.

\subsubsection{Parallelizing M-MIVAs}

As introduced in Sec.~3.4, M-MIVA
performs two key operations at each diffusion step:
1) joint generation of frame and mask sequences, and 2) formation of attention masks for guiding attention layers in subsequent time steps.
When parallelizing multiple $n$ M-MIVAs, 
the video tensor 
is extended to include both the video frames and the subject masks generated by each M-MIVA, represented as
$[I^{1:F}; \{ S^{(j)} \}_{j=1}^n]$.
For clarity, we omit the frame indices $1\text{:}F$ on $S$.
Each module $j$ independently generates its corresponding mask sequence $S^{(j)}$, consistent with the single-motion-pattern setup. However, the construction of attention masks diverges across different attention layer types and is defined as follows:
1) for CFA, CA, and t-SA layers associated with MIVA $j$, each attention mask is computed solely based on the corresponding subject mask sequence $S^{(j)}$, mirroring the single-motion-pattern formulation;
2) for SA layers that are coupled with CFA layers,
attention masks must integrate multiple subject masks from all $n$ M-MIVAs. We define a unified subject mask $S^*$, where each pixel $S^*_{\bp} \in \{0, 1, \dots, n\}$ is given by
\begin{equation}
S^*_\bp = \begin{cases}
\max(\arg \max_{j \in \{1, \dots, n\}}(S^{(j)}_\bp)) & \text{if } \max_{j} S^{(j)}_\bp > 0.5, \\
0  & \text{otherwise.} 
\end{cases}
\end{equation}
$S^*_{\bp}$ stands for the semantic region $\bp$ most likely belongs to, which can be one of the motion subject indices ($1, \dots, n$) or $0$ corresponding to the background.
In case of ambiguity (\ie, overlapping subject masks), we select the largest MIVA index,
assuming that the user specifies the ordering of MIVAs such that a larger index is positioned closer to the top, and a smaller index lies below.
Based on the unified mask sequence $\{S^{*1}, \dots, S^{*F}\}$, we define the unified attention mask $M^*$ with respect to 
the video token pair $(\bp, i)$-$(\bq, j)$ ($\bp,\bq$ are spatial locations and $i,j$ are frame indices here) as:
\begin{equation}
    M^{*(i, j)}_{(\bp,\bq)} = log \left( \mathbf{1} \left( {S^{*i}_{\bp} = S^{*j}_{\bq}} \right) + \epsilon \right)
\end{equation}
where $\epsilon$ is a small positive constant added for numerical stability, and $\mathbf{1}(x)$ denotes an indicator function that returns $1$ if $x$ is true and $0$ otherwise.
$M^*$ is then applied to modulate the SA layers.

The framework also supports the special case of parallelizing MIVAs and M-MIVAs jointly.
In this setup, each vanilla MIVA is assigned a mask sequence matching the background,
that is, the binary mask $\mathbf{1}(S_\bp^* = 0)$,
which serves as the basis for computing its attention masks.
This alignment allows uniform processing within the joint architecture while retaining consistency with M-MIVA.




\subsection{Inference-time Techniques}

Some I2V generation methods employ inference-time tuning-free techniques to enhance the visual quality of output frames \cite{lamp, cil, cinemo, i4vgen, consisti2v}.
These techniques can be categorized into pre-processing and post-processing.
We specify our design choice for inference-time pre- and post-processing of video tensors in this section.
Specially, in M-MIVA inference, the generation of subject mask sequences follows an identical procedure, thereby employing a unified pre-processing pipeline. 
Post-processing is skipped for mask sequences, as M-MIVA outputs video frames only.

\subsubsection{Pre-processing Techniques}
Pre-processing techniques manipulate the initial noise at inference time
$\bx_T$ to replace the original white Gaussian noise.
The underlying intuition is that $\bx_T$ should preserve a small amount of information from $I$ to align with the forward process, 
rather than being left completely random \cite{cil}.
We discover that integrating the shared-noise mechanism by LAMP \cite{lamp} and the DCTInit algorithm by Cinemo \cite{cinemo} results in more desirable frame quality and fidelity to $I$ compared to applying either method alone.
We summarize the pre-processing pipeline in Alg.~\ref{alg:init}.

\begin{algorithm}
\caption{Pre-processing at inference time}
\label{alg:init}
\begin{algorithmic}
\Require VAE encoder $\vae$, the reference image in latent domain $\bx^I = \vae(I)$, low-pass filter $\lpf$, shared-noise coefficient $\alpha$, noise schedule $\{\alpha_T, \sigma_T\}$
Generate        noise signals $\bepsilon^{1, ..., F} \sim N(0, \bI)$\\
\For{$i \in \{2, ..., F\}$}
    \State $\tilde \bepsilon^i \gets \alpha \bepsilon ^1 + (1-\alpha) \epsilon^i$ 
    \Comment{Shared noise by LAMP}
\EndFor

\State $\bx_T \gets \alpha_T \bx^I + \sigma_T \epsilon^1$ \Comment{Forward diffusion}\\
\State $\bX_T \gets \dct_{3D}(\bx_T), \bE \gets \dct_{3D}(\tilde \bepsilon)$\\

\State $\tilde \bX^1 \gets \dct_{3D}(\bx^I)$\\
\For{$i \in \{2, ..., F\}$}
\State $\tilde \bX^i \gets \bX_T \odot \lpf + \bE^i \odot (1-\lpf)$
\Comment{DCTInit; $\odot$ stands for element-wise multiplication}
\EndFor

\State \Return $\idct_{3D}(\tilde \bX)$
\end{algorithmic}
\end{algorithm}

\subsubsection{Post-processing Techniques}
Post-processing techniques manipulate the output of DM, either during the sampling process (\eg, Adaptive Instance Normalization or AdaIN \cite{adain}) or after sampling (\eg, histogram matching with $I$). These techniques are also tuning-free.
In contrast to LAMP which applies AdaIN after each sampling iteration, we observe that applying AdaIN only at the end of the last iteration, right before the VAE decoding step, yields consistently higher video frame quality.
We discard the histogram matching step in LAMP, arguing that an equivalent histogram with $I$ is a poor proxy for temporal coherency, effective only in cases of little dynamics. 



\subsection{Implementation details}

\subsubsection{Implementation of MIVA Submodules}

In total, a MIVA consists of the following learnable parameters:
$\bW_Q, \bW_O$ and the adaptive weighting module of each CFA layer,
matrices $\bA$ and $\bB$ associated with each MLP that replaces CA,
and the adapter attached to each t-SA layer (exclusively for AnimateDiff).
The implementation follows a minimalistic design heavily inspired by LoRA.
Specifically, $\bW_Q$ in CFA layers and the projection matrices in t-SA layers are fine-tuned using respective additive LoRA modules.
Meanwhile, 
$\bW_O$ of CFA is modeled by a low-rank matrix, equivalent to a LoRA attached to zero, and the same applies to the CA-equivalent MLPs.
Empirically, we assign LoRA ranks
of $128$ to CFA layers, $64$ to CA layers, and $32$ to t-SA layers.
This specific configuration has demonstrated consistent convergence across varying motion patterns in the training dataset, with no significant benefit observed when increasing ranks beyond these values.

\subsubsection{MIVA based on AnimateDiff (U-Net)}
Training follows the standard DM training objective, minimizing the denoising loss. We use the Adam optimizer \cite{adam} with a fixed learning rate of $10^{-5}$, excluding the first frame from the loss computation.
Training is conducted over 15K iterations on a single GPU, with a minibatch size of 1. The VRAM usage is 9.8 GB for MIVA and 11.0 GB for M-MIVA.
At inference time, we employ DDIM sampling \cite{ddim} with 50 time steps.
The total number of learnable parameters in MIVA is 35.4M (24.8M in CFA layers, 1.6M in CA layers, and 9.0M in t-SA layers).
This corresponds to approximately 3\% of the base model's 1.29B parameters.
M-MIVA introduces an additional 20.4M learnable parameters for the mask generation stream, increasing the total to roughly 5\%.
We set the classifier-free guidance (CFG \cite{cfg}) scale to $1$ to avoid training and inference of unconditional generation.
  

\subsubsection{MIVA based on Wan-T2V-1.3B (DiT)}
We also train MIVA based on a diffusion transformer (DiT)-based model, Wan2.1-T2V-1.3B \cite{wan}. Bounded by the pre-training configuration of the base model, MIVA is trained to generate videos of resolution $17 \times 832 \times 480$, which differs from the resolution used with AnimateDiff. 
The choice of 17 frames, compared to 16 in AnimateDiff, is dictated by Wan's spatiotemporal VAE, which applies a temporal compression ratio of 4 and requires the input video length to be $4\alpha + 1$.
Here we set $\alpha = 4$, resulting in a 17-frame video compressed into 5 latent frames at $1/8$ resolution.
Training follows the same protocol as AnimateDiff-based MIVA. Pre-trained weights are frozen, and the adapters are optimized over 8K iterations using the Adam optimizer (learning rate $10^{-5}$) with a minibatch size of 1. The VRAM usage during training is 10.4 GB for MIVA and 22.2 GB for M-MIVA.
The MIVA consists of 35.5M trainable parameters (23.6M in CFA layers and 11.9M in CA layers), which is less than 3\% of the base model's 1.3B parameters. M-MIVA introduces an additional 35.5M learnable parameters for the mask generation stream, raising the total to 5.5\%.
All other settings remain consistent with those used in AnimateDiff-based MIVA training.

\subsection{Other Design Options}

We note several MIVA design options that we experimented with but appeared less effective.

\subsubsection{IP-Adapter}
IP-Adapter \cite{ip-adapter}
extends a CA layer by an additional CA term dependent on an input reference image,
enabling a DM to be controlled by not only the text prompt but also by the image.
Originally designed for DM-based image editing, it is later introduced into I2V DMs such as I2V-Adapter \cite{i2v-adapter}, which states that integrating a pre-trained IP-Adapter (termed ``Content-Adapter'' in the paper) improves temporal stability apart from alignment with the reference image.

\begin{figure}[h]
    \centering
    \includegraphics[width=0.45\linewidth]{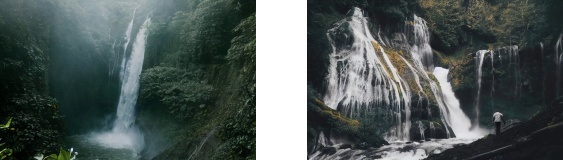}
    \hspace{0.03\linewidth}
    \includegraphics[width=0.45\linewidth]{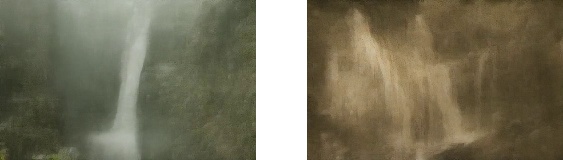} \\
    \parbox{\linewidth}{
        \hspace{0.2\linewidth}
        \small{Input}
        \hspace{0.36\linewidth}
        \small{Frame 16}
    }\\
    \caption{Introducing IP-Adapter into CA layers results in blur and desaturation.}
    \label{fig:ip}
\end{figure}

We follow IP-Adapter by injecting an additional attention operation term for each CA layer, with $K$ and $V$ based on the embedding of $I$, given by the CLIP image encoder \cite{clip} further processed by a pre-trained IP-Adapter. Like I2V-Adapter, we train the learnable adapter weights while freezing the IP-Adapter weights.
However, we observe this harms the resulting quality drastically: as shown in Fig.~\ref{fig:ip}, the generated videos exhibit severe blurring and desaturation artifacts. A possible reason is that the new image-conditioned term may distort the distribution of the CA layer output, requiring more data for the learnable weights to adapt.

\subsubsection{Analytic-Init}
Analytic noise initialization (Analytic-Init)\footnote{Besides Analytic-Init, the paper proposes another technique that is to reduce the initial time step $M$ during sampling, which is originally $T$ in most DMs. For simplicity, we refer to the combination of these two techniques as Analytic-Init.} \cite{cil} is an inference-time pre-processing technique aimed at boosting dynamics without introducing artifacts. We reimplement the method with the hyperparameters $\sigma_p^{2*}=0.566$ and $M \in [0.8, 0.96]$, but do not observe consistent improvements in motion intensity across motion patterns.

\section{Additional Details on Experiments and Evaluation}

\subsection{Our Benchmark Datasets}

The single-motion-pattern benchmark dataset consists of 8 motion patterns following LAMP \cite{lamp}, as listed below:
\begin{itemize}
    \item birds flying ($\caseB$, MIVA prompt being ``birds''),
    \item fireworks ($\caseF$, ``fireworks''),
    \item helicopter hovering ($\caseP$, ``helicopter''),
    \item horse running ($\caseH$, ``horse''),
    \item person playing the guitar ($\caseG$, ``guitar''),
    \item raining ($\caseR$, ``rain''),
    \item person turning to smile ($\caseS$, ``face''), and
    \item waterfall cascading ($\caseW$, ``waterfall'').
\end{itemize}

The 6 motion combinations of the multi-motion-pattern benchmark dataset are based on the 8 motion patterns of LAMP dataset plus ``clouds moving'' ($\caseC$, MIVA prompt being ``clouds'').
They are
\begin{itemize}
    \item person playing the guitar + turning to smile ($\caseG\caseS$),
    \item waterfall cascading + horse running ($\caseW\caseH$),
    \item waterfall cascading + birds flying ($\caseW\caseB$),
    \item waterfall cascading + clouds moving ($\caseW\caseC$),
    \item birds flying + clouds moving ($\caseB\caseC$), and
    \item person turning to smile + clouds moving ($\caseS\caseC$).
\end{itemize}

For simplicity later, we use the corresponding symbol defined above to refer to each motion pattern.

\subsection{Metrics of Quantitative Evaluation}

Below are the objective evaluation metrics used in Sec. 4.
All the metrics originate from VBench \cite{vbench}, except for average flow, which is from EvalCrafter \cite{evalcrafter}.

\subsubsection{Subject Consistency}
evaluates the visual consistency of the foreground subject(s) across frames, ensuring that they maintain a stable appearance.
It is defined as the average cosine similarity (scaled to $[0,100]$) between the DINO \cite{dino} features of consecutive frames to detect any identity shifts in the subject, with high scores indicating a steady visual identity throughout the video.

\subsubsection{Background Consistency}
evaluates the stability of the background scene across frames, ensuring minimal changes as the video progresses.
It is defined as the average cosine similarity (scaled to $[0,100]$) between the CLIP \cite{clip} features of consecutive frames, with high scores reflecting a consistent background.

\subsubsection{Motion Smoothness}
evaluates the realism of motion between frames.
It is calculated through a reconstruction procedure: alternate frames are dropped, predicted with a video frame interpolation mode \cite{amt}, and the mean absolute error (MAE) is computed between the synthesized and original frames. The smoothness score is then the negative MAE normalized to $[0,100]$.
A smaller gap results in higher scores, indicating smoother and more realistic motion.

\subsubsection{Temporal Flickering}
reflects content-independent fluctuations across frames, such as shaky camera motions and variations in lighting and/or exposure.
It is given by the negative MAE between consecutive frames, normalized to $[0,100]$, so that higher scores reflect better temporal stability.


\subsubsection{Aesthetic Quality Loss}
concerns the aesthetic appeal of each frame
measured by the LAION aesthetic predictor \cite{laion}.
Each frame receives a rating between $0$ and $10$,
which is later linearly normalized to $[0,100]$.
Considering that input images vary in perceptual quality, we report the loss of quality compared to $I$ rather than the original scores.
The final score for a video is obtained by averaging the quality loss of all frames.

\subsubsection{Imaging Quality Loss}
addresses low-level image quality, focusing on distortions such as noise, blur, and exposure issues. Each frame's quality is scored by MUSIQ \cite{musiq}, an image quality assessment neural network,
and then normalized to $[0,100]$.
Similar to the aesthetic quality loss, we report the average quality loss of all frames.

\subsubsection{Average Flow}
reflects motion intensity. It is defined by the average optical flow intensity, as provided by RAFT \cite{raft}, between consecutive frames. A higher value indicates greater dynamics within the video.

\subsubsection{Why Are Text Alignment Metrics Not Used?}

Text alignment metrics (\eg, the CLIP alignment score) are widely applied in text-guided generation to assess fidelity to the text prompt. Some I2V methods report the text alignment scores as well.
We argue that these metrics are not suitable for the problem addressed in this paper:
text alignment metrics, which match each frame independently with the text prompts, can only attend to visual content and cannot capture dynamic information, as it can only be discerned by observing multiple frames. Instead, we rely on user study to assess the quality of generated dynamics.


\subsection{User study}

\begin{figure}[h]
    \centering
    \includegraphics[width=\linewidth]{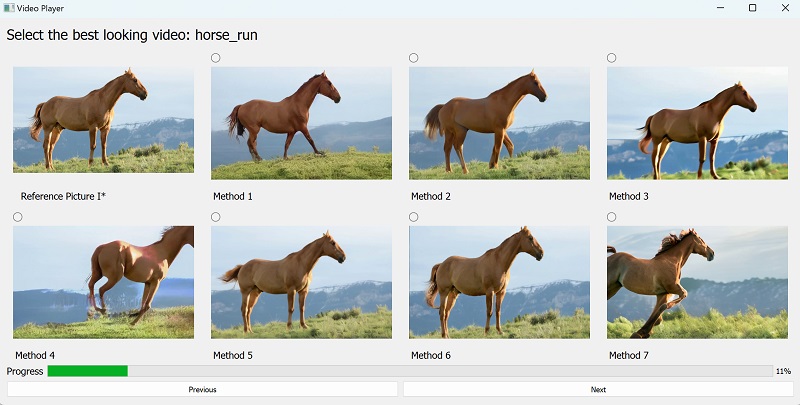}
    \caption{The user interface of the user study program.}
    \label{fig:userstudy}
\end{figure}

We invited 20 viewers to mark the animation results by the studied results using our self-designed program.
All viewers are nonprofessionals in related research fields or the animation industry.
Each viewer was asked to select the best animation in their opinion on 1) 40 random cases from the single-pattern-motion set, 2) 20 random cases from the multi-pattern-motion set, and 3) 15 random cases in a head-on-head comparison between MIVA and M-MIVA on the 3 selected motion patterns ($\caseB$, $\caseF$ and $\caseH$).
As shown in Fig.~\ref{fig:userstudy}, all candidate results are anonymized during the evaluation process.

\section{Additional Results}

\textit{Note: please refer to the \textbf{Project Page} for the result videos in this section, located at}

\url{https://yishaohan.github.io/MIVA-web}






\subsection{Adjusting MIVA Weights in Multi-motion-pattern Animation}


Section 2-2 in the Project Page showcases how adjusting the MIVA weights enables fine-grained control over the intensity of each constituent motion pattern.
Empirically, a total weight of $1$ yields the most stable animation results.

\begin{table*}[t]
    \centering
    \resizebox{0.9\textwidth}{!}{
    \renewcommand{\arraystretch}{1} 
    \begin{tabular}{c|l|ccccccccc}
        \toprule
        \makecell{Training \\ Data} & Method & \makecell{Birds \\$\caseB$} &
        \makecell{Fireworks \\$\caseF$} &
        \makecell{Helicopter \\$\caseP$} &
        \makecell{Horse \\$\caseH$} &
        \makecell{Guitar \\$\caseG$} &
        \makecell{Rain \\$\caseR$} &
        \makecell{Smile\\ $\caseS$} &
        \makecell{Waterfall \\$\caseW$} &
        Overall \\
        \midrule
        \multirow{5}{*}{\makecell{Large-\\scale}} & SVD \cite{svd} &
        3.1 & 0.0 & 5.1 & 6.2 & 3.1 & 3.8 & 4.5 & 0.6 & 3.3 \\
        & I2VGen-XL \cite{i2vgen} & 
        16.2 & \textbf{35.6} & 9.4 & 10.2 & 25.4 & \textbf{29.1} & 3.1 & 11.6 & \underline{17.6}\\
        & DynamiCrafter \cite{dynamicrafter} &
        10.9 & 17.6 & \underline{21.2} & \underline{14.2} & 6.5 & \underline{20.8} & 10.1 & 5.7 & 13.4\\
        & I2V-Adapter \cite{i2v-adapter} &
        1.2 & 15.1 & 2.3 & 1.3 & 3.7 & 10.0 & 2.8 & 6.9 & 5.4\\
        & Cinemo \cite{cinemo} &
        \textbf{30.8} & 5.1 & 8.3 & 0.8 & \textbf{29.8} & 16.1 & \underline{27.4} & \underline{15.5} & 16.7\\
        \midrule
        \multirow{2}{*}{Few-shot} & LAMP \cite{lamp} & 
        15.6 & 5.7 & 7.5 & 1.3 & 5.6 & 13.1 & 19.9 & 4.4 & 9.1 \\
        & MIVA (Ours) &
        \underline{22.2} & \underline{20.9} & \textbf{46.2} & \textbf{66.0} & \underline{25.9} & 7.1 & \textbf{32.2} & \textbf{55.3} & \textbf{34.5}\\
        \bottomrule
    \end{tabular}
    }
    \caption{
    Preference rate (in percentage) per method, per case in our user study on single-motion-pattern animation.
    The best number is highlighted in \textbf{bold}, and the second best is \underline{underlined}.
    } 
    \label{tab:case}
    
\end{table*}
\subsection{Camera Movement by MIVA}

Section 2-3 in the Project Page presents a particular dual-motion-pattern setting where one of the motion patterns originates from the camera itself.
When employing the existing methods, we append additional words, such as ``camera pan right'' and ``camera zoom out'', to the prompt.
However, experimental results reflect that
these models struggle to interpret and implement camera-related cues from text alone, suggesting that they fail to reliably capture motion semantics tied to camera movement.

Inspired by MotionLoRA \cite{animatediff}, we regard camera motions as distinct motion patterns and train dedicated MIVAs to represent them.
Rather than collecting video datasets with explicit camera movement (e.g., LAMP), we generate synthetic training data using a curated set of 20 high-resolution images of natural scenes. For each image, we apply random cropping followed by resizing or translation to simulate zoom and pan movements. This procedure yields five videos per image, resulting in a dataset of 100 samples for each camera motion type.
At inference time, camera motion MIVAs and subject motion MIVAs are integrated via weighted summation, consistent with the fusion strategy detailed earlier. Qualitative results demonstrate that the inclusion of camera-specific MIVAs enables precise and reliable control over camera dynamics, overcoming the limitations of text-only prompting.






\subsection{Wan-based MIVA}

Section 4 in the Project Page exhibits animations generated by Wan-based MIVAs, demonstrating their compatibility with DiTs.
Wan-based MIVAs demonstrate consistent performance with their AnimateDiff-based counterparts across a variety of motion patterns (e.g., $\caseF$, $\caseG$, $\caseP$, $\caseR$, $\caseW$).
However, they exhibit limitations when handling more subtle motions, such as facial expressions, bird wing flapping, and horse leg movement, corresponding to $\caseS$, $\caseB$, and $\caseH$.
We speculate that this performance gap arises from the compactness of the latent video tensor: Wan VAE compresses the original 17-frame input into 5 latent frames at 1/8 resolution, which may be insufficient for capturing fine-grained dynamics.
We leave the exploration on DiT-based MIVAs in larger-scale settings (\eg, for generating longer videos), for future work.


\section{Discussions}





\subsection{Scope of Modular I2V Setting}
Modular I2V decomposes dynamics into one or more atomic motion patterns, implicitly assuming that each pattern behaves independently.
While effective for disentangled motion representations, this assumption can break down when modeling more intricate scenarios such as:
\begin{itemize}
    \item Intricate inter-subject interactions, \eg, a person pulling a fish out of water with a fishing rod.
    \item Multi-action behavior by a single subject, 
    \eg, a person showing varied facial expressions while talking on the phone, where the animation of face can be a challenge.
\end{itemize}


Additionally, ideal parallelism for MIVA is achieved when
motion patterns are non-overlapping,
allowing Eq.~\eqref{eq:supp} to aggregate the residuals from each MIVA without mutual interference.
Otherwise, overlapping regions risk inducing crosstalk effect \cite{oa}, where multiple MIVAs interfere and generate entangled or conflicting dynamics.
For instance, in the $\caseB \caseC$ case (see Section 2-1 in the Project Page), the cloud motion overrides the dynamics of the birds.









\subsection{Failure Cases}

The per-case performance evaluation (Tab.~\ref{tab:case}) reflects that MIVA is well received in most motion patterns, with the notable exception of rain motion.
While existing methods, especially those trained on large-scale datasets, have relatively consistent performance in animating rain, our method suffers from insufficient dynamics in some cases. 
We find that MIVA can actually capture the dynamic information of rain; yet the pre-processing steps are the bottleneck, which enforce excessive alignment with the first frame, although they benefit animation quality for most motion patterns.
This highlights the need for further exploration of inference-time strategies to better accommodate globally distributed and fine-grained motions such as rain.

We also note some noticeable visual artifacts that are common in I2V DMs, including MIVA:
\begin{itemize}
    \item Distortions to human face. Animation models often modify the appearance of subjects, and such modifications are mostly sensible, especially in natural scenes. However, modifications to human face are more challenging due to the subtle, salient details that are more noticeable to human eyes. Even a slight distortion can lead to greater perceived unnaturalness than some natural objects, such as a waterfall. It can be seen from the animation results that all studied methods suffer from unnatural distortions to the human face to varying degrees.
    Some papers address the face distortion issue by embedding additional person-identity-related information into the DM, \eg, \cite{id-animator, photomaker}.
    \item Poorer quality for small-scale objects. Among the studied motion patterns, birds flying generally exhibits the poorest animation quality. Upon close inspection, viewers can often see birds vanishing, blinking or spawning from nothing. A likely underlying reason is that the dimensionality of spatially (and also temporally in Wan-based MIVAs) downscaled video tensors in the latent domain is too low to capture the subtle details of small objects precisely.
\end{itemize}

\subsection{M-MIVA}

In Section 4.3, we demonstrate the performance gains achieved by M-MIVA across
three selected motion patterns ($\caseB, \caseF, \caseH$).
However, M-MIVA shows limited improvement over the remaining five motion patterns in the LAMP dataset:
\begin{itemize}
    \item Helicopter $\caseP$: Due to the subject's rigid structure and clear separation from the background, mask guidance offers minimal enhancement beyond MIVA's baseline effectiveness. 
    \item Raining $\caseR$: The high density and rapid movement of raindrops make accurate mask generation infeasible, diminishing the utility of mask-based guidance. 
    \item Guitar playing $\caseG$, Turning to smile $\caseS$ and Waterfall $\caseW$: These cases exhibit minimal subject silhouette motion, which poorly reflects the internal dynamics of the subject. Consequently, mask guidance provides negligible benefit in capturing meaningful motion cues. 
\end{itemize}
These observations suggest that the effectiveness of mask guidance is sensitive to motion characteristics, and its effectiveness hinges on how well the mask can represent underlying dynamic patterns.




\section{Broader Impacts}
As a low-cost approach that can be easily deployed on personal computers, MIVA might be utilized by malicious users to create disturbing or unsafe animations. Additionally, MIVA may inherit the biases and flaws present in the base DM.
Addressing these concerns, we highlight the importance of adhering to ethical guidelines when using the proposed method and advocate for necessary supervision to ensure ethical use.















\section{Acknowledgment}

We sincerely thank the contributions and comments from 
\textit{Daixin Tian} (University of Toronto),
\textit{Yilun Jiang} (University of Toronto),
and \textit{Zihuan Jiang} (University of Toronto).


\end{document}